# DeepTCM1.0: A Multi-Expert AI Agent for Deciphering Mechanisms of Chinese Herbal Formulae Based on General Large Language Models

Wenxin DUAN[1,2†], Hanwei WANG[1,2†], Zhongying PENG[1,2], Zhonghua LU[2], Jiayi AN[3], Fan SONG[2*] & Yong LIANG[2*]

†Wenxin Duan and Hanwei Wang have contributed equally to this work.

*Correspondence: Fan Song, fansong616@163.com;

Yong Liang, liangy02@pcl.ac.cn.

[1] Guangzhou University of Chinese Medicine, 232 Outer Ring East Road, Guang Zhou 510006, China.

[2] Chinese Medicine Guangdong Laboratory (Hengqin), 2522 Huandao North Road, Zhu Hai 519031, China.

[3] Changzhi Medical College, 161 Jiefang East Road, Luzhou District, Changzhi 046000, China.

**Abstract:**

**Background:** Mechanistic elucidation of traditional Chinese medicine (TCM) compound formulas remains a central challenge in the modernization of TCM. Conventional approaches, including data mining and network pharmacology, are insufficient for achieving deep integration between classical TCM theory and modern scientific research. In addition, direct question-answering using general-purpose artificial intelligence large language models is limited by inadequate adaptation to TCM theoretical frameworks and susceptibility to reasoning hallucinations. Consequently, there is an urgent need to develop intelligent analytical methods aligned with the holistic principles of TCM.

**Objective:** To establish a multi-expert intelligent agent framework integrating classical TCM theory with modern life sciences, thereby enabling systematic and interpretable mechanistic analysis of TCM compound formulas, with Guizhi Decoction serving as a representative validation case.

**Methods:** The DeepTCM1.0 framework was constructed based on the general-purpose large language model DeepSeek V3.2. It adopts a three-tier collaborative architecture and a three-round iterative quality-control workflow, simulating the collaborative analytical process of 11 interdisciplinary intelligent agents. The framework was applied to the mechanistic interpretation of Guizhi Decoction from the dual perspectives of classical traditional Chinese medicine theory and modern scientific research. Framework performance was comprehensively evaluated through double-blind five-dimensional scoring, intraclass correlation coefficient (ICC) reliability testing, Mann-Whitney U tests, and effect size analysis. The evaluation employed four independent large language models as evaluators, each conducting five rounds of repeated scoring on five anonymized reports, resulting in a total of 100 independent scoring assessments.

**Results:** The framework systematically elucidated the relationship between the classical traditional Chinese medicine pathogenesis of “disharmony between ying and wei” and modern neuro-immune-metabolic regulatory networks. It further proposed three innovative mechanistic hypotheses, addressing several longstanding controversies, including the identification of pharmacodynamic material foundations and the linkage between macroscopic syndrome manifestations and microscopic molecular mechanisms. Evaluation results demonstrated robust reliability and validity. Intra-rater reliability was favorable, with ICC values for individual evaluators ranging from 0.789 to 0.867, while inter-rater reliability was excellent (Krippendorff’s $\alpha = 0.868$). One-way analysis of variance revealed highly significant intergroup differences ($F = 66.969$, $P < 0.001$), with an effect size of $\eta^2 = 0.738$, indicating excellent discriminative validity. Furthermore, a highly significant difference was observed between the multi-agent group and the single-LLM group ($U = 191.00$, $Z = -5.403$, $P < 0.001$), with an effect size of $r = 0.5403$, reaching the threshold for a large effect, thereby confirming that the overall performance of DeepTCM1.0 was significantly superior to that of single general-purpose large language models. Radar chart visualization further showed that DeepTCM1.0 achieved near-perfect scores across all evaluation dimensions (4.8-5.0 points) and exhibited the highest scoring consistency among all tested models. In addition, the framework offers practical advantages including lightweight deployment, zero fine-tuning requirements, and high reproducibility.

**Conclusion:** DeepTCM1.0 enables deep integration between classical TCM theory and modern scientific research, providing a standardized and intelligent paradigm for mechanistic analysis of TCM compound formulas. This framework may substantially advance the modernization of TCM and interdisciplinary research integrating traditional Chinese and Western medicine.

## 1.Introduction

Traditional Chinese Medicine (TCM), as a valuable part of Chinese cultural heritage, adheres to the key principles such as 'pattern differentiation and treatment, holistic regulation, and formula compatibility.' Classical prescriptions (jingfang) are employed as the primary therapeutic carriers. Owing to their characteristics of multi-component, multi-target, and multi-pathway, these formulations have exhibited unique advantages in the prevention and treatment of complex diseases. However, the scientific elucidation of the molecular mechanisms underlying herbal compound prescriptions, as well as the integration of traditional theories with modern empirical evidence, has consistently been identified as a central challenge in the modernization of TCM research[1-3].

Over the past decades, advances in biology, chemistry, medicine, pharmacy, and information science have been leveraged by interdisciplinary researchers to elucidate the TCM mechanisms from various perspectives [4,5].In particular, within the domain of information science, methodological approaches to investigating TCM mechanisms have evolved from data mining to network pharmacology, and more recently to deep learning models [6-8]. Data mining methods, grounded in association rule learning and frequency analysis, have been using techniques such as Support Vector Machines (SVM), Random Forests (RF), and Logistic Regression (LR) to identify core herb combinations within TCM formulae, thereby providing actionable references for personalized clinical treatment and promoting the advancement of TCM research and clinical practice [8].

However, traditional data mining methods alone are insufficient to reveal the complex synergistic relationships among the components of herbal formulas. They also fail to explore molecular regulatory networks. Consequently, such approaches remain largely confined to descriptive associations and phenomenological observations[8]. With the advancement of high-throughput technologies, the emergence of network pharmacology has facilitated the transition of TCM research into the realm of biomolecular networks. By employing core methodologies such as target prediction, network topology analysis, and pathway enrichment, studies at the levels of genes and proteins have achieved a paradigm shift from "single-target investigation" to "network-based regulation"[13, 23]. Building upon this foundation, regulatory networks centered on "component-target-pathway" relationships have been constructed to uncover the underlying principles within TCM. When combined with phytochemical analysis and various experimental assays, this approach has become a fundamental methodology for elucidating the holistic mechanisms of TCM [9,23,24].

Nevertheless, traditional network pharmacology methods exhibit clear limitations. Available datasets are often fragmented and update infrequently, thereby restricting the reliability of analyses and predictions. Furthermore, network models are predominantly static, limiting their capacity to capture the dynamic and temporal characteristics of herbal formula efficacy. Additionally, the integration of multi-scale information across molecules, cells, and tissues levels remains insufficient, highlighting a critical methodological gap [10,25].

On the other hand, with the widespread application of molecular docking technologies, specialized docking software such as Molecular Operating Environment (MOE) [47], LeadIT [48], and Genetic Optimization for Ligand Docking (GOLD) [49] have been developed. These tools have enabled researchers to analyze the mechanisms of TCM using methodologies using conceptual frameworks derived from conventional drug discovery. By optimizing computational parameters, the interaction modes of compounds with specific targets can be simulated, thereby facilitating the screening and ranking of potential drug molecules. Although molecular docking technology has been applied to the field of TCM research for several years, its core design and technical logic primarily revolve around the development of small chemical molecules. Consequently, it is inherently difficult to align with the fundamental principles of TCM, including pattern differentiation-based treatment, holistic regulation, and formula compatibility. In addition, limitations such as insufficient functional compatibility, high operational costs, and low research efficiency further restrict its applicability. As a result, molecular docking approaches have not become a mainstream methodology for mechanistic studies in TCM [11,18].

Over the past three years, the rapid development of artificial intelligence (AI) technology has been propelled by a synergistic combination of algorithmic innovation, breakthroughs in computational hardware, and the substantial growth of high-quality research data [52]. Notably, large language models (LLMs) based on the Transformer architecture, such as GPT, Llama, and DeepSeek, have demonstrated strong strong capabilities in text comprehension, knowledge integration, and logical reasoning. These features have enabled broad applications in the life sciences [15,22,50,52]. Representative examples include the ophthalmology model EYE-Llama [19], a serological testing model for hepatitis [12], and the Rheumatoid Arthritis model Hengqin-RA-v1 [54]. In these studies, distilled versions of LLMs were fine-tuned using high-quality domain-specific datasets, thereby enhancing domain knowledge representation and logical reasoning performance in specialized medical fields. Collectively, these findings provide strong evidence that LLMs can be deeply adapted to biomedical research scenarios and can deliver precise support for scientific investigation in specific sub-fields.

Currently, general-purpose LLMs large models are undergoing rapid development. Their breadth of knowledge coverage, integrated reasoning capacity, and cross-domain adaptability have been continuously enhanced, enabling the synthesis of core knowledge across disciplines and demonstrating substantial potential for scientific research applications. The significant advancements spurred by intelligent agent systems based on large models, have yielded notable progress in frontier life sciences. Two studies published in Nature provide compelling evidence. First, the DeepRare system built on DeepSeek V3, has achieved breakthroughs in precise diagnosis for rare diseases [21]. Second, a large model-driven virtual laboratory, supported by a multi-agent framework, has successfully completed a fully closed-loop research tasks from the design to validation of novel nanoantibodies [53]. These accomplishments collectively demonstrate that general LLMs-driven multi-agent frameworks have emerged as a research hotspot and have gained broad recognition in the international academic community.

TCM constitutes a complex scientific system, for which the development of intelligent agent systems is urgently required to extend the capabilities of general purpose large language models such as GPT and DeepSeek, for innovative research [14,36]. Owing to advantages including multi-expert autonomous exploration, collaborative interaction, and emergent reasoning, agent-based systems exhibit a high degree of compatibility with the systemic characteristics of TCM, particularly its holistic perspective, pattern differentiation–based treatment, and formula compatibility. Furthermore, the mechanisms of consensus-seeking, evidence-based querying, and critical reflection among multiple expert agents have been shown to effectively reduce hallucination phenomena in LLMs, thereby offering substantial potential for mechanistic studies in TCM [13,16,17,20], as illustrated in Figure 1.

Therefore, this study proposes a multi-expert intelligent agent framework, termed DeepTCM1.0, based on the deep exploration of TCM mechanisms via general LLMs. The open-source domestic general LLM DeepSeek V3.2 was employed as the central "brain" of AI agents. Through the design of specialized and precisely defined expert roles, a progressive and interactive task decomposition strategy was implemented, alongside a three-stage hierarchical iterative analysis process incorporating quality control and validation. This framework aims to achieve deep interdisciplinary integration of classic TCM theories and modern life sciences, constructs a traceable research evidence chain, and establishes an intelligent analytical method aligned with core TCM principles, namely pattern differentiation-based treatment, holistic regulation, and compound prescription.

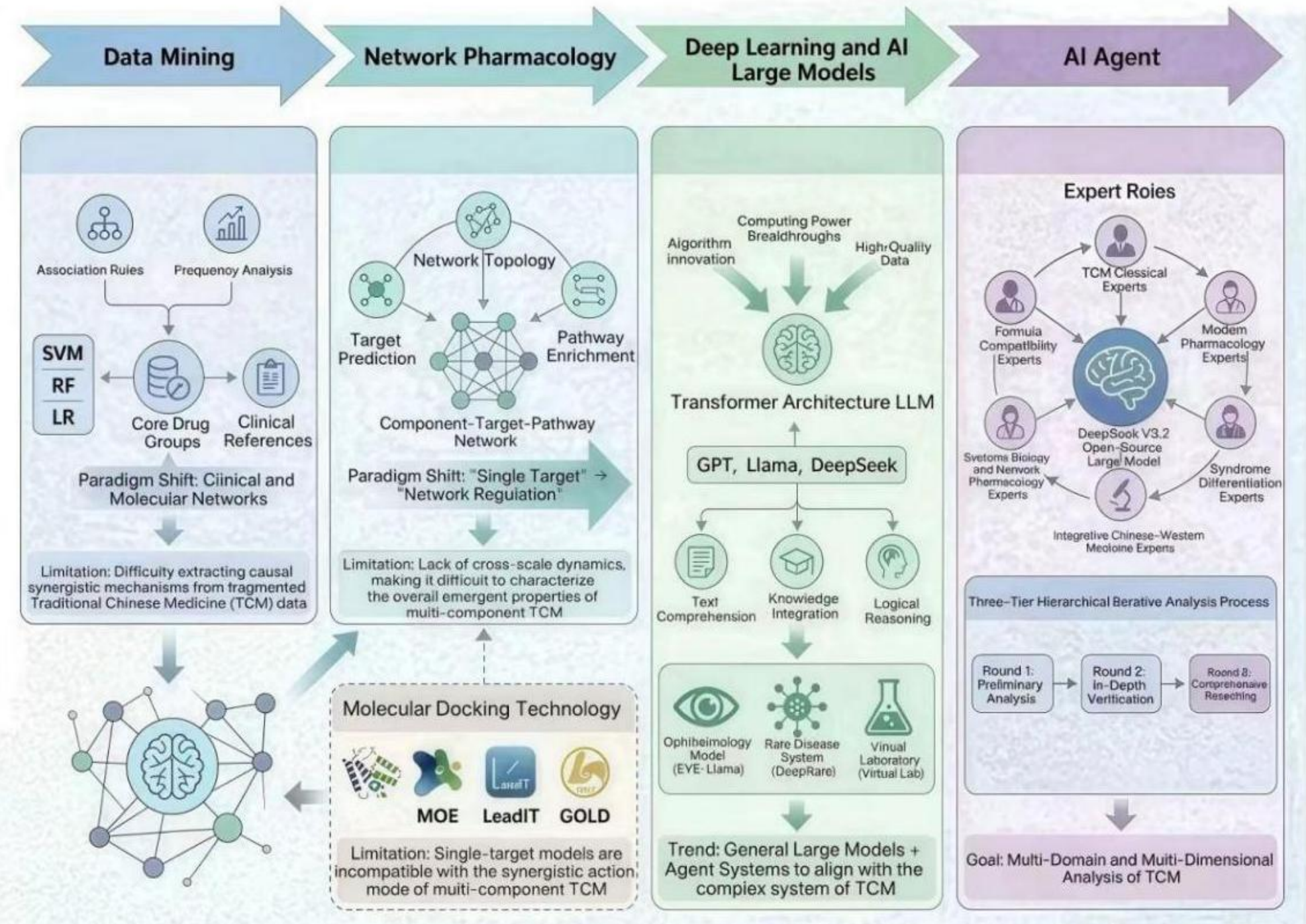

**Figure 1. The Evolution of Information Science Technologies in Traditional Chinese Medicine Mechanism Research**

DeepTCM1.0 simulates the collaborative analyses of experts across various domains, including classical TCM, formula compatibility, modern pharmacology, network pharmacology, molecular biology, and molecular docking, to comprehensively elucidate the mechanisms of herbal formulas from multiple perspectives. In addition, dynamic mappings between abstract core concepts of TCM and measurable indicators in modern biology are established, thereby enabling the efficient decomposition and execution of complex interdisciplinary research tasks.

Using Guizhi Decoction(桂枝汤) as a representative case, the effectiveness of DeepTCM1.0 in mechanistic analysis was systematically demonstrated. This study aims to provide a reproducible technical framework for classic formula research and syndrome mechanisms elucidation, thereby facilitating the transition of TCM from experience-based knowledge transmission to intelligence-driven scientific innovation.

## 2 Methods

### 2.1 Overall Design Strategy

The core design principle of DeepTCM1.0, the multi-expert agent collaboration framework developed in this study, lies in leveraging the 'plug-and-play' capability of LLMs to simulate multidisciplinary collaboration mechanisms through expert role configuration, thereby enabling the comprehensive analysis of the mechanisms underlying traditional Chinese medicine (TCM) formulae.

The framework design adheres to the following four principles: ① TCM Theory-Oriented. Centered on the core logic of "Holistic perspective, Pattern Differentiation-based treatment, and hierarchical principle of Monarch–Minister–Assistant–Courier," the configuration of agent expert roles and the discussion process are guided by classical TCM theories; ② Multidisciplinary Synergy. Integrating perspectives from diverse fields such as classical TCM, formula compatibility, phytochemistry, network pharmacology, and molecular biology to avoid the limitations of single-discipline approaches; ③ Lightweight Implementation. Intrinsic multidisciplinary knowledge embedded within the general-purpose model DeepSeek V3.2 was extensively exploited to stimulate integrative reasoning capabilities. Notably, no additional knowledge graphs were constructed, and no further model fine-tuning was performed; ④ Interpretability Priority. Through multi-round iterative discussions and an evidence-based reasoning process, the framework ensures that analytical conclusions are clearly traceable, with logic that progresses layer by layer and remains self-consistent.

Based on the four principles above, DeepTCM1.0 was designed as a three-tier collaborative framework: the first tier is the task decomposition and team formation layer; the second tier is the multi-expert collaborative reasoning and emergent insight layer; and the third tier is the integrative validation and output layer.

## 2.2 Three-Tiered Collaboration Framework

DeepTCM1.0 adopts a three-tier modular closed-loop operational architecture, as illustrated in Figure 2: (a) Task Deconstruction and Team Formation Layer: Human research questions are received and parsed through the agent interface, after which the Chief Scientist decomposes the task and strategically assembles a dedicated interdisciplinary expert team. (b) Multi-Expert Collaborative Reasoning and Emergence Layer: Fixed-domain experts, dynamically recruited specialists, and critic agents engage in multiple rounds of cross-disciplinary reasoning through a collaborative inference engine, substantially reducing AI hallucinations and reasoning bias. (c) Integration, Refinement, and Research Proposal Output Layer: Through iterative multi-agent review, quality control, and optimization, the system ultimately generates logically coherent scientific hypotheses and standardized research protocols with traceable evidence chains.

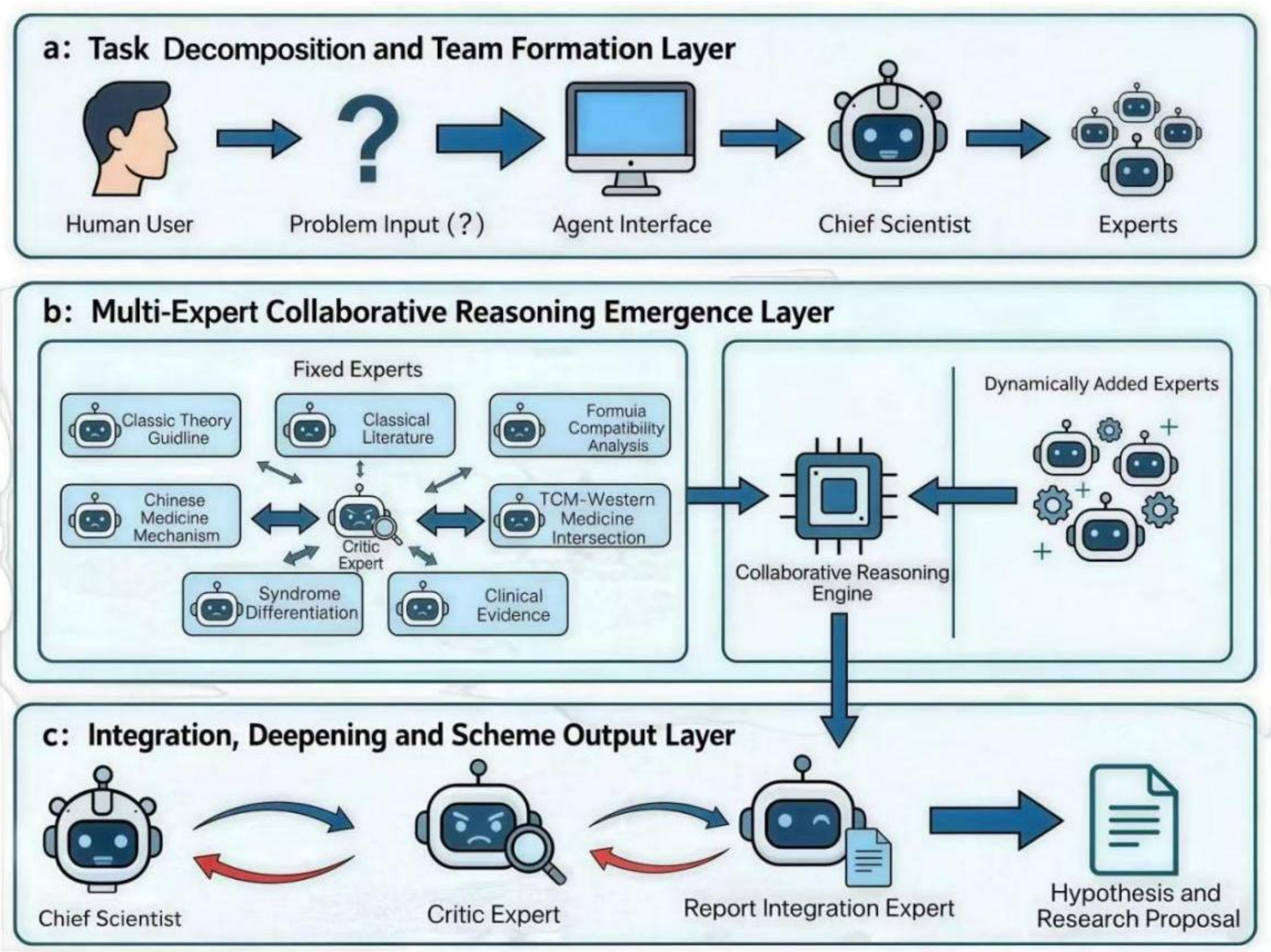

**Figure 2. Three-Tier Collaborative Operational Architecture of the DeepTCM1.0 Multi-Expert Agent Framework**

### 2.2.1 First Tier: Task Decomposition and Team Formation

This tier serves as the core decision-making and control hub for the entire research process, responsible for planning research directions and designing analytical pathways. It encompasses two core modules: Principal Investigator and Recruiter. The primary function of the Principal

Investigator is to decompose the complex task of elucidating the mechanisms of traditional Chinese medicine (TCM) compound prescriptions into standardized, multidimensional professional sub-tasks. The architectural design and operational logic strictly adhere to the research philosophy of TCM-led interdisciplinary collaboration.

The Recruiter assembles an expert team based on these sub-tasks, integrating both fixed experts (whose domains and role prompts are predefined by human experts) and dynamic experts (whose domains and role prompts are defined by the Principal Investigator and recruitment officer according to task requirements). This joint configuration ensures comprehensive disciplinary coverage without perspective gaps. Such a collaborative structure not only guarantees the integrity and professionalism of TCM formula research but also leverages the extensibility and forward-looking capacity of modern scientific approaches, thereby supporting the systematic integration of traditional TCM with cutting-edge technologies.

**Operational Mechanism:** The Principal Investigator receives user input (e.g., "Comprehensively analyze the efficacy mechanism of Guizhi Decoction from classical TCM theory and modern scientific perspectives") and performs task decomposition according to predefined research principles. This includes delineating specialized domains, such as classical theoretical tracing, formula compatibility analysis, and pharmacodynamic material basis with pharmacokinetic evaluation, resulting in six core sub-tasks.

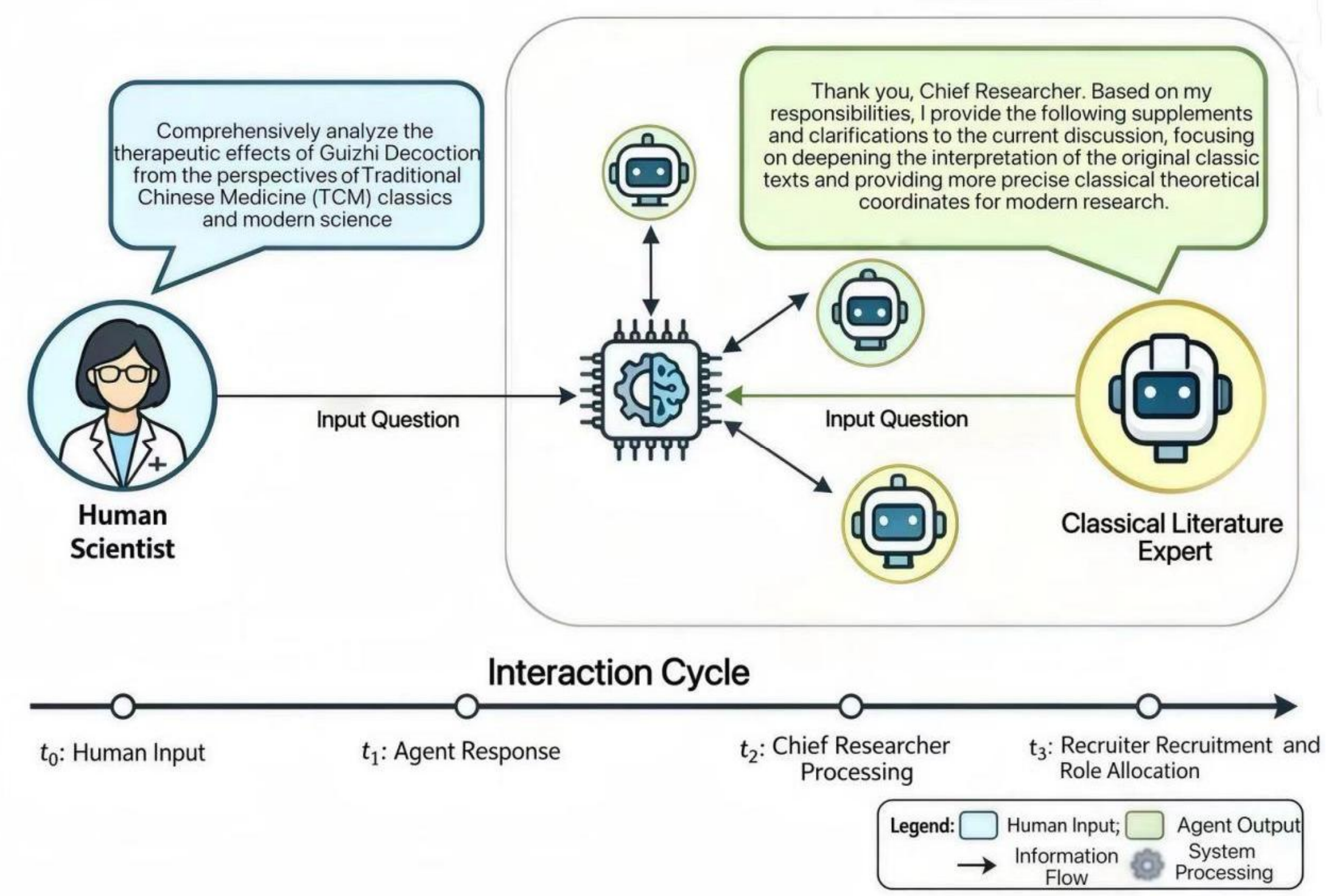

**Figure 3. Agent Role-Playing Framework: Interdisciplinary Dialogue and Knowledge Integration**

Based on this decomposition, the recruitment officer constructs a joint expert team. For example, fixed experts may include specialists in classical TCM literature, formula compatibility, and TCM mechanisms, while dynamic experts may cover fields such as immunology and neuroendocrinology, metabolomics and microbiomics, and related disciplines. The resulting team spans key domains, including core TCM theory, formula systems, modern medicine, pharmacology, and integrative Chinese–Western medicine.

Each expert agent was equipped with a dedicated, standardized role prompt that precisely defined its knowledge background, core responsibilities, and criteria for professional interpretation, thereby ensuring a focused analytical perspective and compliance with disciplinary standards.

For example, the prompt assigned to the Classical Literature Expert was specified as follows: "You are an expert in classical TCM literature, with comprehensive expertise in canonical texts and historical commentaries, and an integrated perspective bridging classical theory and modern microscopic research. Your core tasks include providing original classical passages relevant to the topic, specifying their sources, offering modern-language interpretations, and comparing interpretations from historical commentators. You are required to analyze differences among scholarly traditions and the evolution of theoretical frameworks, thereby establishing a classical foundation for modern research. Furthermore, you should elucidate the correspondence between classical theory and modern molecular mechanisms, syndrome biomarkers, and other microscopic research findings, construct a conceptual bridge between classical knowledge and contemporary science, clarify the guiding significance of classical theory, and propose research pathways linking ancient and modern perspectives. In addition, you should identify hypotheses derived from classical texts that remain unverified and propose corresponding experimental directions. The standards for interpretation must ensure citation accuracy, clear source tracing, integration of classical and modern perspectives, and strict adherence to the core principles of classical theory", as illustrated in Figure 3.

This tier is grounded in core TCM principles, including the holistic perspective, the sovereign–minister–assistant–courier hierarchy, and pattern differentiation–based treatment (bianzheng lunzhi), while extending its reasoning framework to encompass integrative Chinese–Western medicine, macro–micro alignment, and interior–exterior coherence. It establishes a foundational research paradigm characterized by "TCM-guided, multidisciplinary collaboration, and specialized division of labor."

By design, this tier prevents issues such as topic deviation, disorganized task allocation, capability gaps, and logical discontinuities. It provides a robust foundation for subsequent iterative analyses in the following tiers. Guided by TCM theory in both agent role configuration and task decomposition, and implemented through a hybrid structure of fixed and dynamically recruited experts, the framework effectively operationalizes interdisciplinary integration.

Notably, no additional database construction or model fine-tuning is required; instead, the extensive pretrained knowledge embedded in the general-purpose model DeepSeek V3.2 is fully utilized.

Furthermore, the standardized task decomposition and expert collaboration design provide structural support for comprehensive recording of multi-round iterative discussions and reasoning processes. This ensures that the conclusions derived from TCM formula mechanism analysis are fully traceable, logically coherent, and progressively structured, thereby adhering to the principle of prioritizing interpretability.

### 2.2.2 Second Tier: Multi-Expert Collaborative Reasoning Layer

This layer serves as the intermediate execution unit within the three-tier framework, linking top-level task deployment with the expert team. Its primary responsibility is to coordinate multidisciplinary experts in conducting in-depth, multi-perspective analyses centered on a unified research theme.

Within this layer, a critic role is specifically introduced, specializing in reverse-evidence-based reflective reasoning. This role focuses on systematically examining the integrity of logical arguments, the authenticity and sufficiency of research evidence, and the compatibility between classical theoretical frameworks and modern scientific findings. By identifying conflicts in viewpoints, research blind spots, and omissions in expression, the critic continuously advances the refinement of research perspectives through sustained questioning and targeted optimization recommendations, thereby providing essential quality control support for the rigor of research outcomes. The bidirectional reasoning and constructive debate between the critic and the expert team not only significantly enhance the capacity of general-purpose large language models to generate new discoveries and mechanistic insights through emergent reasoning, but also markedly reduce the probability of hallucinations.

This layer employs a “three-round iterative” discussion mechanism, coordinated by the Principal Investigator, who maintains the central research trajectory and enforces discussion standards. The Principal Investigator organizes expert analyses, synthesizes viewpoints, focuses on key controversies, integrates macro- and micro-level reasoning, corrects inferential gaps, and promotes interdisciplinary consensus formation.

The three iterative rounds are structured as follows.
First, the independent interpretation phase: experts from each discipline conduct independent analyses without external influence, ensuring that viewpoints remain unbiased and grounded in their respective professional perspectives. Each expert produces an initial analytical report that is objective and aligned with both TCM theoretical principles and modern biomedical knowledge. The critic then conducts reverse reasoning and evidence-based questioning for each initial report, while leveraging large language models to collect and organize relevant supporting literature.

Second, the integration and deepening phase: under the coordination of the Principal Investigator, experts and the critic collectively advance the discussion following the sequence of “TCM theory first, modern pharmacology second, and cross-validation third.” During this stage, insights are systematically organized, consensus areas are identified, and key controversies are delineated, thereby consolidating the analytical foundation.

Third, the optimization and enhancement phase: the expert team and the critic jointly address the controversies identified in the second round through targeted inquiries, supplementary evidence, and multidisciplinary in-depth analysis. Diverse forms of evidence are incorporated to correct earlier deviations, while AI-assisted reasoning within each disciplinary domain is applied to generate novel research hypotheses addressing core issues and filling existing research gaps.

### 2.2.3 Third Tier: Integrative Deepening and Solution Output Layer

This tier represents the final closed-loop stage and outcome delivery phase of the entire workflow. A report integration specialist is introduced at this stage, serving as the central entity responsible for comprehensive information integration and the systematic construction of research outputs. This role consolidates key research elements generated throughout the three iterative rounds, including core viewpoints, consensus conclusions, points of controversy, and hypotheses pending validation. These elements are then systematically examined and organized according to a standardized structural framework of “classical theory—macroscopic logic—microscopic mechanisms—clinical application,” ensuring that all research information remains complete, traceable, and clearly sourced.

Through intensive collaborative interaction among the Principal Investigator, the critic, and the report integration specialist, this tier accomplishes the integrated verification, structured presentation, and standardized delivery of research outcomes, ultimately producing a compound-mechanism analysis report characterized by logical completeness, comprehensive evidence support, and rigorous professional standards. Unlike the second tier, in which expert teams conduct multi-perspective analyses around specific research topics, this tier places greater emphasis on synthesizing and refining existing analytical conclusions, conducting evidence-based quality control, and elevating findings into a coherent and systematic framework.

Within this collaborative structure, the three roles perform distinct yet complementary functions, working in a progressive sequence defined as “strategic coordination—evidence-based scrutiny—integrative optimization.” The Principal Investigator establishes direction and makes strategic decisions, the critic conducts continuous quality assessment and corrective review, and the report integration specialist formalizes and standardizes the final documentation. Through this mechanism, the diverse viewpoints generated in the second tier are synthesized into scientifically coherent and verifiable hypotheses, along with a complete research plan. The final outputs consist of implementable and testable compound-mechanism research schemes and consensus conclusions, thereby achieving standardized and systematic integration of multi-expert deliberations and completing the full analytical closed loop for compound-mechanism research.

### 2.3 Technical Implementation

#### 2.3.1 Development Environment

This study was developed using Python 3.10. Core modules included an OpenAI-compatible API endpoint (for invoking DeepSeek API), python-dotenv (for environment variable management), and JSON (for structured data processing), etc.. The runtime environment consisted of the Windows 11 operating system, with 16 GB memory and an Intel Core i7 CPU.

#### 2.3.2 Implementation of AI Agents

An object-oriented programming (OOP) approach was adopted. A unified base class, Agent, was defined, incorporating attributes such as name (role identifier), role_description (role-specific instruction prompts), and expertise_area (specialized field). Core methods included analyze() (for independent analytical tasks) and discuss() (for multi-agent discussion participation). Each expert agents was customized via refined role prompt engineering. Prompt design followed a four-stage structure: Role Definition, Knowledge Background, Task Objectives, and analytical Boundaries.

#### 2.3.3 Multi-Round Discussion Process Control

A Discussion Manager class was designed and implemented to orchestrate the multi-round discussion process, which maintained a complete dialogue history, and recorded speeches from each round. After each discussion round, the Critic was invoked to perform quality control (QC) evaluation. The system supports breakpoint resumption and intermediate result caching to ensure stability for long-duration analytical tasks. To guarantee full reproducibility of the analytical process, all prompts, intermediate outputs, and final reports were locally persisted in JSON format.

#### 2.3.4 Distinctive Advantages

The core advantages of the technical framework in this research include: (1) Zero additional data cost was achieved by fully leveraging the pre-trained knowledge embedded in general LLMs, without requiring domain-specific database construction; (2) Zero fine-tuning cost was maintained, as domain adaptation was realized exclusively through prompt engineering and expert role simulation; (3) Full reproducibility was ensured as all prompts and process designs are shared, facilitating replication and adaptation of the framework for other classic TCM formula studies.

### 2.4 Case Analysis Design

To validate the effectiveness of DeepTCM 1.0 agent framework, a comprehensive analysis was conducted on the therapeutic mechanism of Guizhi Decoction as a representative case. Guizhi Decoction, originally recorded in the Shanghan Lun, is recognized as a foundational classical prescription. Its core efficacy is described as harmonizing Yin and Yang (yin–yang denotes the dynamic balance of functional activities in the body), and regulating Ying and Wei (nutritive and defensive qi). It embodies core TCM theories including disharmony of Nutrient and Defensive Qi, imbalance of Yin and Yang, and exterior-resolution and interior-regulation, thereby serving as a paradigmatic example of TCM holistic practice.

Theoretically, as a classic formula, Guizhi Decoction has numerous derivatives, so its modern mechanism elucidation provides a representative model for related classical formulae. Practically, Guizhi Decoction is widely used clinically, with extensive existing evidence from network pharmacology and metabolomics studies [27,31,32], including metabolic regulation [27,46],neuro-immune modulation [28],and antipyretic mechanisms [29,30]. These multidimensional data sources align well with the interdisciplinary collaborative capabilities of multi-expert agent systems, enabling validation of the DeepTCM 1.0 in terms of knowledge integration, mechanism discovery, and evidence traceability.

The analysis task is defined as: "To systematically elucidate the therapeutic mechanism of Guizhi Decoction from classical TCM theory and modern scientific perspectives". The expert team was assembled through a hybrid strategy combining fixed core experts and dynamically augmented specialists. The disciplinary coverage included core TCM theory, formula composition, modern medicine, pharmacology, and integrative Chinese–Western medicine, ensuring comprehensive expertise and adaptability to diverse analytical tasks.

The expected output structure includes: classical theory interpretation (original textual analysis of Guizhi Decoction in the Shanghan Lun, correspondence between formula and syndrome, and principles of formulation); compatibility mechanism analysis (the roles of sovereign–minister–assistant–courier, herb–herb interactions, and compatibility characteristics); material basis analysis (major active constituents and their source medicinal materials); network regulatory mechanisms (core targets, key pathways, and multi-target synergistic networks); integrative conclusions (the mapping relationship between TCM theory and modern mechanisms, and the core mode of action); and validation recommendations and research protocols (key targets and pathways for subsequent experimental verification, as well as proposed research plans).

## 3 Results

### 3.1 Agent System Execution Process

Following the three-layer collaboration framework and the three-round iteration mechanism, the DeepTCM 1.0 agent system completed a full-process analysis of the efficacy mechanism of Guizhi Decoction.

#### 3.1.1 Task Deconstruction and Team Formation

Upon receiving the task, "comprehensively analyzing the therapeutic mechanism of Guizhi Decoction from both classical TCM theory and modern science perspectives", the Principal Investigator decomposed it into six sub-tasks: (1) Traceability of Classical Theory; (2) Analysis of Formula Compatibility; (3) Identification of bioactive material basis and pharmacokinetic analysis; (4) Exploration of Network Regulation Mechanisms; (5) Analysis of Metabolomics and Microbiomics; (6) Integration of Clinical Evidence.

The system dynamically assembled a collaborative team consisting of 11 experts based on subtask requirements. The team included 7 fixed core experts, covering fields such as classical theory architecture, TCM ancient texts, formula compatibility, Chinese medicine mechanisms, Chinese-Western interdisciplinary studies, syndrome differentiation, and clinical evidence. In addtion, in conjunction with the Principal Investigator, 4 specialized experts were incorporated according to task demands. These specialists were customized with standardized role prompts tailored to distinct disciplinary domains, including immunology and neuroendocrinology, metabolomics and microbiomics, systems biology and network pharmacology, and pharmacokinetics and pharmaceutical analysis.

The entire process was conducted without manual intervention, ensuring comprehensive disciplinary coverage while enhancing the depth of interdisciplinary collaboration.

#### 3.1.2 Multi-Expert Collaborative Reasoning

**First round (independent interpretation):** all eleven experts independently elaborated their viewpoints based on their respective professional perspectives. After passing critical inquiry and quality control audits, an initial analysis report was generated, establishing a multi-dimensional evidence foundation. The core conclusions of each expert are detailed in Table 1.

**Table 1. Core Conclusions of Each Expert During the First Round of Independent Interpretation**

| Expert Role | Core Work Content | Independent Interpretation Conclusions |
|---|---|---|
| Classic Theory Guidline Expert | This role adheres to the core principles of classical TCM theory, defines the essential pathogenesis of “disharmony between ying and wei (nutritive and defensive qi),” and ensures that the analytical logic remains aligned with the classical framework. | First, the fundamental nature of ying–wei disharmony is characterized by dysregulated outward dispersion of wei qi and a relative insufficiency of ying yin, which must be distinguished from pure deficiency syndromes. The theoretical foundation is established in Article 58 of the Shanghan Lun, which states that “when yin and yang achieve self-regulation, recovery will naturally occur.”<br><br>Second, the ultimate therapeutic objective of Guizhi Decoction is to restore the dynamic balance of yin and yang, rather than to merely exert anti-inflammatory or antipyretic effects.<br><br>Third, the spleen and stomach are regarded as the source of the transformation and generation of ying–wei. The “middle energizer (zhongjiao) functional unit,” composed of ginger, jujube, and licorice, together with the procedure of “taking warm porridge and maintaining warmth” (cuozhou wenfu), constitutes an indispensable component of therapeutic efficacy and should not be omitted. |
| Classical Literature | This role focuses on textual scholarship of Guizhi Decoction, | First, the core provisions are derived from Articles 12, 13, |

| Expert Role | Core Work Content | Independent Interpretation Conclusions |
| --- | --- | --- |
| Expert | including clause-based indications, dosage ratios, and interpretations by historical commentators, thereby establishing a documentary foundation that bridges classical theory with modern research. | 53, and 54 of the Shanghan Lun. The 1:1 ratio of Guizhi (Cinnamomi Ramulus) to Shaoyao (Paeoniae Radix) is critical for achieving a balance between dispersion and astringency. Cheng Wuji, in Annotated Treatise on Cold Damage, explicitly emphasized its principle of "combining dispersion and restraint."<br><br>Second, the practice of "taking warm, diluted porridge" (cuo re xi zhou) is regarded as a form of "natural nutritional adjuvant," which replenishes ying yin and enhances drug absorption. Xu Lingtai highlighted that it "supplements grain qi to augment the efficacy of the formula."<br><br>Third, the "moderate pulse (mai huan)" primarily reflects insufficiency of nutritive qi and reduced vascular tension, rather than heart rate. Fang Youzhi, in Clause Differentiation of the Treatise on Cold Damage, clarified its opposing relationship with the "tight pulse (mai jin)." |
| Formula Compatibility Analysis Expert | This role focuses on elucidating the sovereign–minister–assistant–courier (jun–chen–zuo–shi) structure and compatibility principles of Guizhi Decoction, | First, the formula can be divided into two functional modules: the "ying–wei unit" (Guizhi + Shaoyao) and the "middle energizer (zhongjiao) unit" |

| Expert Role | Core Work Content | Independent Interpretation Conclusions |
|---|---|---|
| | linking macroscopic therapeutic effects with microscopic molecular interactions. | (ginger + jujube + licorice), with nearly equivalent dosage proportions, reflecting the principle of “treating disease by addressing its root cause.”<br><br>Second, the herb pair Guizhi–Shaoyao exerts synergistic immunoregulatory effects through the “TRPV1–NF-κB pathway.” Specifically, Guizhi, characterized by its pungent and dispersing properties, activates TRPV1 receptors, whereas Shaoyao, with sour and astringent properties, inhibits the NF-κB pathway, thereby achieving immune–inflammatory balance.<br><br>Third, the compatibility principle of “pungent–sweet generating yang” (Guizhi + licorice) corresponds to β2-adrenergic receptor (β2-AR) activation, while “sour–sweet generating yin” (Shaoyao + licorice) corresponds to the inhibition of pro-inflammatory cytokines, including IL-6 and TNF-α. The synergistic effects of these compatibilities should be further validated through formula decomposition studies. |
| Chinese Medicine Mechanism Expert | This role interprets the active constituents, molecular targets, and signaling pathways of individual herbs based on classical properties | First, the Principal component of Guizhi (Cinnamomi Ramulus), cinnamaldehyde, acts as a TRPV1 agonist that |

| Expert Role | Core Work Content | Independent Interpretation Conclusions |
|---|---|---|
| | (nature and flavor), and establishes a mapping framework of “property–component–target.” | promotes the release of calcitonin gene-related peptide (CGRP), thereby exerting antipyretic and vasodilatory effects [J Nat Prod, 2005]. The key component of Shaoyao (Paeoniae Radix), paeoniflorin, functions as a TLR4/NF-κB pathway inhibitor, producing anti-inflammatory effects and contributing to yin preservation [Int Immunopharmacol, 2010].<br><br>Second, glycyrrhetinic acid, a metabolite of glycyrrhizic acid derived from licorice, inhibits 11β-hydroxysteroid dehydrogenase type 2 (11β-HSD2), thereby enhancing the local effects of cortisol [Mol Cell Endocrinol, 2014].<br><br>Third, the overall formula exerts synergistic multi-target regulation through the “TRPV1–NF-κB–HPA axis.” In vitro experiments should be designed to correspond with in vivo exposure concentrations to ensure translational relevance. |
| Syndrome Differentiation Expert | This role analyzes the four diagnostic features of the “Taiyang wind-strike pattern with exterior deficiency” and establishes a linkage framework between macroscopic syndromes and microscopic biomarkers. | First, the core syndrome is characterized by “fever, spontaneous sweating, aversion to wind, and a floating–moderate pulse.” These manifestations correspond to specific microscopic pathophysiological changes: |

| Expert Role | Core Work Content | Independent Interpretation Conclusions |
| --- | --- | --- |
| | | fever is associated with elevated hypothalamic prostaglandin E2 (PGE2); spontaneous sweating corresponds to abnormal expression of aquaporin 5 (AQP5) in sweat glands; aversion to wind is related to increased sensitivity of cutaneous TRPM8 receptors; and the floating–moderate pulse reflects reduced vascular tone.<br><br>Second, the microbiological basis of spontaneous sweating is defined as dysfunction of the cholinergic nerve–sweat gland axis, and assessment of acetylcholine receptor activity is therefore required.<br><br>Third, syndrome differentiation should be supported by microscopic indicators. Compared with the Taiyang cold-damage excess pattern, the distinguishing features lie in sweat gland function (spontaneous sweating versus anhidrosis) and vascular tone (moderate versus tight pulse). |
| Integrative Chinese–Western Medicine Expert | This role integrates insights from modern biomedical research to construct a contemporary pathological model of "disharmony between ying and wei (nutritive and defensive qi)," thereby bridging classical theory with modern scientific logic. | First, a "three-tier, dual-axis, one-foundation" model is proposed. This framework comprises the core regulatory layer (neuro–endocrine–immune, NEI network), the core effector layer (vascular–microcirculatory system), and |

| Expert Role | Core Work Content | Independent Interpretation Conclusions |
| --- | --- | --- |
| | | the foundational support layer (gut–metabolic system). The thermoregulatory axis and the barrier immunity axis are identified as the central axes.<br><br>Second, ying–wei disharmony is conceptualized as an acute maladaptation of the NEI network, manifested by disruption of inflammatory balance, autonomic nervous system dysregulation, and abnormal rhythmicity of the hypothalamic–pituitary–adrenal (HPA) axis.<br><br>Third, Guizhi Decoction exerts its therapeutic effects through the synergistic modulation of the “neural–immune–endocrine–metabolic” network, thereby restoring systemic homeostatic resilience. |
| TCM Clinical Evidence Expert | This role evaluates clinical evidence for Guizhi Decoction and designs empirical protocols aligned with real-world clinical practice. | First, the current level of clinical evidence is graded as B–C, indicating moderate-quality support for its efficacy in treating the exterior deficiency pattern of common cold. Reported outcomes include reductions in pro-inflammatory cytokines such as IL-6 and TNF-α [J Ethnopharmacol, 2014].<br><br>Second, clinical studies should strictly include patients meeting the “four-symptom |

| Expert Role | Core Work Content | Independent Interpretation Conclusions |
| --- | --- | --- |
| | | combination” (fever, spontaneous sweating, aversion to wind, and floating–moderate pulse), while excluding contraindicated populations, such as habitual alcohol users, patients with a tight pulse, and those without sweating.<br><br>Third, the procedure of “taking warm porridge and maintaining warmth” (cuozhou wenfu) should be standardized as follows: administration of 200 mL millet porridge at 60–70 °C, followed by thermal maintenance using a constant-temperature electric blanket for 1 h. A control group without porridge administration should be established to evaluate its synergistic effect. |
| Systems Biology and Network Pharmacology Expert | This role constructs a “component–target–pathway” network for Guizhi Decoction and quantitatively evaluates multi-target synergistic effects. | First, the predicted core targets include AKT1, TNF, IL6, and CHRM3, while the Principal pathways involve NF-κB signaling, cholinergic synapses, and hypothalamic–pituitary–adrenal (HPA) axis–related pathways.<br><br>Second, network topology analysis indicates that the constituent compounds of Guizhi Decoction are significantly enriched in functional modules associated with immune regulation, neural |

| Expert Role | Core Work Content | Independent Interpretation Conclusions |
|---|---|---|
| | | regulation, and metabolic regulation, thereby reflecting the network-based characteristics of "harmonization" in TCM.<br><br>Third, a critical bottleneck lies in the lack of data on "syndrome-specific targets." To address this limitation, integration with clinical multi-omics data is required to supplement and refine network nodes. |
| Immunology and Neuroendocrinology Expert | This role defines a model of neuro–endocrine–immune (NEI) network imbalance and proposes a core panel of immunological and neuroendocrine biomarkers. | First, acute maladaptation of the NEI network is characterized by an abnormal TNF-α/IL-10 ratio, reduced high-frequency (HF) power of heart rate variability (HRV), and disrupted cortisol circadian rhythm.<br><br>Second, the core biomarker panel includes the TNF-α/IL-10 ratio, salivary cortisol rhythmicity, HRV indices (HF/LF ratio), and salivary secretory immunoglobulin A (sIgA).<br><br>Third, Guizhi Decoction may restore immune–neuroendocrine balance by modulating macrophage polarization, promoting regulatory T cell (Treg) differentiation, and stabilizing hypothalamic– |

| Expert Role | Core Work Content | Independent Interpretation Conclusions |
| --- | --- | --- |
| | | pituitary–adrenal (HPA) axis function. |
| Metabolomics and Microbiomics Expert | This role analyzes the effects of Guizhi Decoction on host metabolic profiles and gut microbiota, and proposes a “microbiota–metabolism–immunity” regulatory axis. | First, under conditions of ying–wei disharmony (nutritive–defensive imbalance), metabolic alterations include enhanced glycolysis, increased degradation of branched-chain amino acids (BCAAs), and a reduced abundance of butyrate-producing bacteria, such as Faecalibacterium prausnitzii.<br><br>Second, Guizhi Decoction modulates gut microbiota composition through polysaccharides derived from ginger, jujube, and licorice, thereby increasing the production of short-chain fatty acids (SCFAs), including butyrate and propionate, which mediate immune–metabolic regulation.<br><br>Third, the administration of warm porridge (cuozhou) provides readily fermentable substrates, promoting SCFA generation and synergistically improving metabolic homeostasis in conjunction with the pharmacological effects of the formula. |
| Pharmacokinetics and | This role elucidates the in vivo fate of active constituents in Guizhi | First, cinnamaldehyde exhibits an oral bioavailability of less |

| Expert Role | Core Work Content | Independent Interpretation Conclusions |
|---|---|---|
| Pharmaceutical Analysis Expert | Decoction and defines a dynamic ensemble of “prototype compounds–metabolites–microbiota-derived products.” | than 1% and predominantly exerts its effects in vivo in the form of its metabolite, cinnamic acid. Paeoniflorin is hydrolyzed by gut microbiota into more active metabolites, thereby enhancing its pharmacological activity.<br><br>Second, ginger, jujube, and licorice have been shown to inhibit intestinal CYP3A4 activity, thereby increasing the bioavailability of active constituents derived from Guizhi (Cinnamomi Ramulus) and Shaoyao (Paeoniae Radix).<br><br>Third, the in vivo exposure profile of bioactive substances should be characterized using ultra-performance liquid chromatography–quadrupole time-of-flight mass spectrometry (UPLC–Q-TOF/MS) [25]. This approach, in combination with pharmacokinetic–pharmacodynamic (PK–PD) modeling, enables the identification and validation of key effect-mediating substances. |

**Second Round (Integration and Deepening):** In this stage, the Principal Investigator coordinated the entire second-round discussion. The discussion agenda and core tasks were clearly defined, focusing on issues identified in the first round, including logical discontinuities, weak macro–micro linkages, and conflicting viewpoints. Three key modules—rectification and verification, conflict integration, and framework construction—were established.

Task standardization was reinforced by requiring experts to submit structured linkage argument tables, with standardized expression and proper citation practices to enhance methodological rigor at the source. During conflict integration, two central issues were emphasized: the delineation of the material basis of efficacy and the construction of macro–micro bridging logic. Relevant experts were guided to engage in cross-disciplinary discussions.

Targeted improvement strategies were proposed to address research gaps identified by the critic agent, thereby strengthening the analytical framework. Meanwhile, a unified consensus was achieved: the in vivo bioactive basis of Guizhi Decoction was defined as a dynamic ensemble of prototype compounds, metabolites, and microbiota-derived products; and a three-tier linkage model of "classical syndromes–intermediate physiological phenotypes–molecular targets" was established, enabling the formation of a complete logical closed loop.

The critic, serving as an independent role dedicated to questioning and evidence-based inquiry, is responsible for engaging individually with each expert to address issues such as conflicting viewpoints, missing citations, and non-standardized expressions. The critic also conducts real-time evaluations of newly emerging issues and hypotheses during the discussion process and provides targeted recommendations for optimization.

The illustrative example is as follows:

**Critic Agent (To the Classical Literature Expert):**

① Regarding the issue of non-standard citations in the first round: specific article numbers and sections from the Shanghan Lun were required to ensure traceability and verifiability, avoiding vague references.

② Regarding the ambiguous interpretation of "Floating-Slow Pulse": it was clarified that "moderate" refers to reduced vascular tension rather than a slowed pulse rate, and supplementation with interpretations from historical commentators was requested, along with connections to modern quantitative indicators.

③ Regarding non-standard hypothesis expression: the proposed association between wei qi and short-chain fatty acids (SCFAs) lacked direct classical support and was required to be reformulated as a scientific hypothesis with a defined validation strategy.

④ Rectification requirements: all revisions were to be systematically verified for accuracy, standardization, and consistency with classical sources, ensuring operational feasibility of proposed linkages.

**Classical Literature Expert (Responding to Critic):**

The constructive comments and rigorous evaluation provided by the critic are gratefully acknowledged. The identified issues have been carefully revised in accordance with the specified requirements, and a standardized response is presented as follows.

First, citation standards have been completed and refined. Specific textual references from the Shanghan Lun have been explicitly supplemented. The core indications of Guizhi Decoction are traced to Articles 12 and 13 in the section Differentiation and Treatment of Taiyang Disease, while the pathogenesis of "disharmony between ying and wei" is referenced to Articles 53 and 54. The procedural method of "taking warm porridge and maintaining warmth" (cuozhou wenfu) is identified in the post-prescription commentary of Article 12. All citations have been standardized to ensure completeness and traceability.

Second, the interpretation of the "floating–moderate pulse (mai fu huan)" has been standardized. The clarification that a "moderate pulse" reflects a relaxed pulse condition has been adopted, and classical commentaries by Cheng Wuji, Fang Youzhi, and Ke Yunbo have been incorporated. The concept of "moderate pulse" is thus defined as reduced vascular tone and diminished elasticity, corresponding to a relative insufficiency of ying yin. In addition, modern quantitative indicators, including pulse wave velocity (PWV) and vascular tone indices, have been proposed to bridge classical theory with contemporary biomedical measurement.

Third, the formulation of hypotheses has been standardized. The proposed association between wei qi and short-chain fatty acids (SCFAs) has been explicitly framed as a scientific hypothesis. It is clarified that no direct evidence currently supports a linkage between the TCM concept of wei qi and SCFAs. This hypothesis is derived from the gut microbiota–immune axis theory and requires experimental validation. A potential validation strategy is proposed, involving comparative analysis of SCFA levels and immune-related biomarkers between patients with ying–wei disharmony and healthy controls.

Fourth, the argumentation has been further strengthened. On the basis of the revisions, additional interpretations from classical commentators regarding "cuozhou wenfu" and the compatibility of ginger, jujube, and licorice have been incorporated. These classical insights have been integrated with modern research directions, including animal model design, identification of bioactive material bases, and data integration frameworks. In this manner, quality control revisions have been closely aligned with research design, ensuring that classical theory serves as a foundational anchor throughout the analytical process.

**Round 3 (Optimization and Refinement):** The Principal Investigator coordinated and led the third round of discussion, focusing on problems raised by the Critic in the second round, including explicit articulation of logical linkages, terminology standardization, and literature supplementation. The optimization strategy emphasized three key elements: hypothesis falsifiability, data linkage, and adherence to theoretical constraints.

Experts were guided to address deficiencies such as implicit macro–micro connections and non-operationalized concepts by providing further clarification and methodological refinement.

Core TCM concepts, including “superficial deficiency (biaoxu),” “taking warm porridge and maintaining warmth,” and the “moderate pulse,” were systematically re-examined through classical interpretation, modern translation, and mechanistic analysis, thereby promoting deeper integration between traditional theory and research design.

Five core theoretical constraints proposed by classical theory experts were incorporated into a dynamic network integration model and treated as fixed conditions for empirical design. All supplementary insights, including literature evidence, operational recommendations, and mechanistic interpretations, were synthesized into the final research framework. A multi-level hypothesis falsification system and a bidirectional statistical analysis strategy were established. Furthermore, a time-series multivariable causal association discovery (TMCAD) workflow was designed as a key technical pathway to validate the novel proposition that “bioactive substances improve symptoms by driving network-level changes.”

Finally, a comprehensive assurance system integrating “top-level theoretical constraints–falsifiability criteria–data-driven validation” was constructed, thereby strengthening the logical coherence and the robustness of macro–micro integration.

**Execution Example: Statement by the Classical Literature Expert (Optimized Following Critic Feedback)**

The rigorous evaluation and constructive feedback provided by the critic are sincerely appreciated. In strict accordance with the required standards, classical citations have been refined, hypothesis statements have been corrected, and classical interpretations have been recalibrated. The optimized conclusions from this round are presented as follows.

First, precise tracing of original classical sources has been achieved. The core indications and the mechanism of harmonizing ying–wei (nutritive and defensive qi) for Guizhi Decoction have been accurately traced to the section Differentiation and Treatment of Taiyang Disease in the Shanghan Lun. Key provisions from Articles 12, 2, 53, and 54 have been comprehensively annotated, clearly specifying the formula indications, pathogenesis, and diagnostic criteria. All citations are consistent with the original text and free from subjective bias.

Second, alignment between classical theory and the core consensus has been established. The integrated formulation strategy and combined use of medicinal substances and dietary therapy in Guizhi Decoction correspond to the concept of a dynamic ensemble of bioactive substances composed of prototype compounds, metabolites, and microbiota-derived products. Furthermore, the three-tier bridging framework of “classical syndromes–intermediate physiological phenotypes–molecular targets” is consistent with the TCM diagnostic logic of “inferring internal conditions from external manifestations” (si wai chuai nei).

Third, unverified hypotheses have been standardized. The proposed association between the porridge-administration method of Guizhi Decoction (cuozhou) and gut microbiota–derived metabolites has been explicitly defined as a scientific hypothesis. This hypothesis is derived from the theory presented in the Lingshu, specifically the chapter Ying Wei Sheng Hui, and remains to be experimentally validated. A feasible validation strategy involves measuring relevant metabolic and microbiota-derived products.

Fourth, supporting evidence from classical commentators has been incorporated. Interpretations from classical scholars such as Xu Lingtai and Ke Yunbo have been integrated to substantiate that the fundamental therapeutic principle of Guizhi Decoction is the harmonization of ying–wei and the restoration of yin–yang self-regulation. These classical perspectives further support the consensus regarding the "multi-component dynamic ensemble of bioactive substances."

#### 3.1.3 Integration, Verification and Output

The Principal Investigator, the critic, and the report integration specialist participate throughout the entire discussion process and ultimately present the consensus findings in a structured format, producing a comprehensive set of conclusions on the mechanistic interpretation of Guizhi Decoction, along with mechanistic hypotheses and corresponding research plans.

### 3.2 Core Consensus on the Mechanism of Action of Guizhi Tang

After three rounds of iterative collaborative discussion, the DeepTCM 1.0 agent system reached the following consensus on the efficacy mechanism of Guizhi Decoction.

#### 3.2.1 Core Positioning at the Classical Theory Level

The fundamental pathogenesis addressed by Guizhi Decoction is identified as "disharmony between ying and wei (nutritive and defensive qi, representing internal nourishment and external defense functions)." This condition is characterized by dysregulated outward dispersion of *wei-qi* and a relative deficiency of *ying-yin*. The statement in Article 12 of the Shanghan Lun—"in Taiyang wind strike, yang is floating and yin is weak"—establishes the theoretical basis for harmonizing ying and wei. The classical 1:1 ratio of Guizhi (Cinnamomi Ramulus) to Shaoyao (Paeoniae Radix) is critical for achieving a balance between dispersion and astringency. Deviation from this ratio may result in excessive diaphoresis or excessive yin-constraining effects, thereby impairing the harmonizing function.

In addition, the "middle energizer (zhongjiao) functional unit," composed of ginger, jujube, and licorice, together with the procedure of "taking warm porridge and maintaining warmth" (cuozhou wenfu), constitutes an integral component of therapeutic efficacy. This reflects the classical concept that "the spleen and stomach serve as the source of ying–wei transformation and generation."

#### 3.2.2 Mechanism Analysis at the Modern Science Level

Based on the integration of multidisciplinary evidence, the mechanism of Guizhi Decoction can be summarized as a “three-tier network harmonization” model.

First, at the core regulatory layer (neuro–endocrine–immune, NEI network), Guizhi Decoction exerts synergistic multi-component effects to restore system resilience. Cinnamaldehyde activates TRPV1 receptors to initiate peripheral signaling; paeoniflorin inhibits the NF-κB pathway to modulate immune–inflammatory balance; and glycyrrhetinic acid regulates the hypothalamic–pituitary–adrenal (HPA) axis to stabilize neuroendocrine rhythms. These effects collectively enable cross-system integration of neural, immune, and endocrine regulation.

Second, at the core effector layer (vascular–microcirculatory system), abnormal sweating is improved through modulation of the cholinergic nerve–sweat gland axis; floating–moderate pulse manifestations are alleviated through regulation of vascular tone; and antipyretic effects are achieved via downregulation of hypothalamic prostaglandin E2 (PGE2).

Third, at the foundational support layer (gut–metabolic system), polysaccharides derived from ginger, jujube, and licorice act as prebiotic substrates to promote the proliferation of butyrate-producing bacteria, such as Faecalibacterium prausnitzii, thereby increasing the production of short-chain fatty acids (SCFAs), including butyrate and propionate. These metabolites contribute to metabolic homeostasis by regulating intestinal barrier function and immune cell phenotypes. The intake of warm porridge further provides fermentable substrates, synergistically enhancing metabolic regulation.

#### 3.2.3 Material Basis and In Vivo Process

The bioactive basis of Guizhi Decoction in vivo is characterized as a dynamic ensemble comprising prototype compounds, metabolites, and microbiota-derived products. Cinnamaldehyde exhibits an oral bioavailability of less than 1% and primarily exerts its effects in the form of its metabolite, cinnamic acid, which locally activates TRPV1 receptors. Paeoniflorin undergoes hydrolysis by gut microbiota into more active metabolites, thereby enhancing its biological activity. Glycyrrhizic acid is converted into glycyrrhetinic acid, which predominantly mediates anti-inflammatory effects.

Furthermore, ginger, jujube, and licorice have been shown to inhibit intestinal CYP3A4 activity, thereby increasing the bioavailability of active constituents derived from Guizhi and Shaoyao. This finding highlights the synergistic enhancement effects achieved through formula compatibility.

### 3.3 Resolution of Core Controversies and Blind Spot Remediation

Through multiple rounds of iterative discussion, the DeepTCM 1.0 agent system effectively resolved several key controversies.

#### 3.3.1 Definition of Pharmacodynamic Material Basis

An initial debate existed between a “prototype compound–dominant” perspective and a “metabolite-dominant” perspective. Following cross-disciplinary deliberation between pharmacokinetics experts and TCM mechanism specialists, a consensus was reached: the therapeutic effects of Guizhi Decoction arise from a dynamic ensemble of prototype compounds, metabolites, and microbiota-derived products, with different forms predominating at distinct stages.

### 3.3.2 Construction of Macro-Micro Connection Logic

A major challenge initially lay in the difficulty of mapping syndrome-based indicators to microscopic biomarkers. Through collaboration between Chinese-Western interdisciplinary experts and syndrome differentiation experts, a three-level connection framework, "Classic Syndrome - Intermediate Physiological Phenotype - Microscopic Molecule", was established. Specifically, four key syndrome elements, Fever, Sweating, Aversion to Wind, and Floating-Slow Pulse, were mapped to measurable physiological indicators, including hypothalamic prostaglandin E2 (PGE2), sweat gland aquaporin 5 (AQP5), cutaneous TRPM8 sensitivity, and vascular tone, as well as molecular targets such as NF-κB, CHRM3 and TRPV1.

### 3.3.3 Standardization of Research Scheme

Divergent experimental strategies were initially proposed for validating the biological basis of the “Guizhi Decoction pattern.” After three rounds of cross-disciplinary discussion among 11 experts, two approaches were established: (i) a composite animal model combining cold exposure (“wind–cold stimulation”) with low-dose lipopolysaccharide (LPS), and (ii) a deep phenotyping clinical study. In both approaches, the classical procedure of “taking warm porridge and maintaining warmth” (cuozhou wenfu) was standardized as follows: administration of 200 mL millet porridge at 60–70 °C, followed by maintenance of a thermally stable environment at 32–34 °C for 1 h.

In the animal model, a multi-level evaluation system was employed, encompassing behavioral outcomes, neuro–endocrine–immune (NEI) indices, vascular microcirculation, and gut microbiota–metabolite profiles. In the clinical study, high temporal resolution sampling (T0–T5) was implemented, and bidirectional regulatory effects were quantified using a deviation convergence (Z-score) model in combination with mixed-effects modeling.

## 3.4 Innovative Mechanism Hypotheses and Research Plans

Based on the above consensus, the DeepTCM 1.0 agent system proposed the following testable mechanistic hypotheses s and corresponding validation research plans.

**Hypothesis 1: Three-Layer Network Harmonization Hypothesis and Its Research Plan**

Guizhi Decoction achieves ying–wei (nutritive–defensive qi) harmonization through cross-level coordination among the “core regulatory layer (NEI network),” the “core effector layer

(vascular–microcirculatory system),” and the “foundational support layer (gut–metabolic system).”

Specifically, a group of key bioactive compounds, including cinnamaldehyde and paeoniflorin, initiates peripheral signaling via activation of cutaneous TRPV1 receptors. This is followed by modulation of immune–inflammatory balance through the NF-κB pathway, and subsequent stabilization of neuroendocrine rhythms via the hypothalamic–pituitary–adrenal (HPA) axis. Concurrently, gut microbiota–derived short-chain fatty acids (SCFAs) optimize metabolic homeostasis, forming a closed-loop regulatory system characterized by “peripheral initiation-central integration-foundational support,” as illustrated in Figure 4.

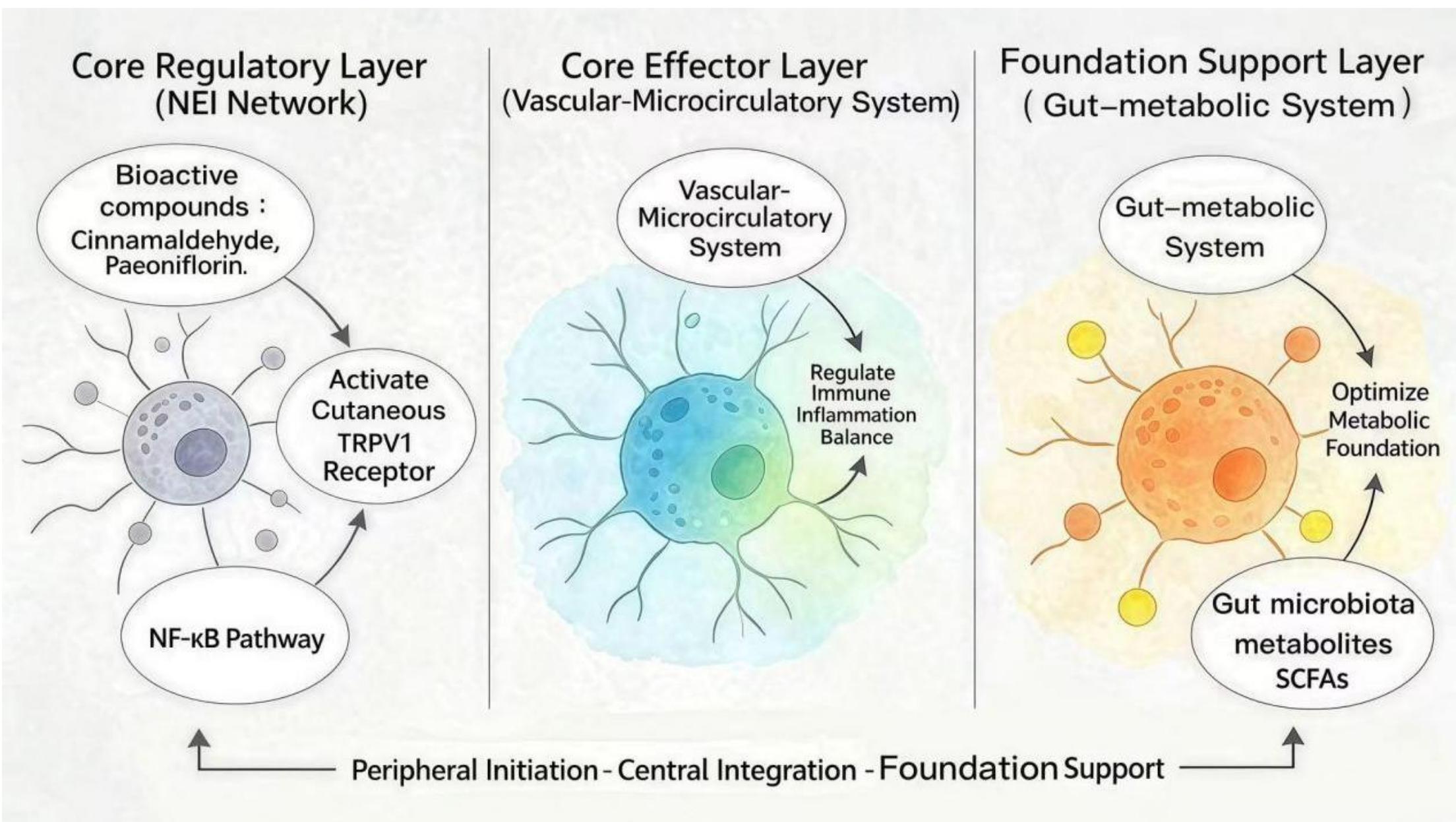

**Figure 4. The Three-Layer Network Harmonization Hypothesis**

**Hypothesis 2: Compatibility-Metabolism-Microbiota Synergy Hypothesis and Its Research Plan**

The core 1:1 compatibility of Guizhi (Cinnamomi Ramulus) and Shaoyao (Paeoniae Radix) exerts immunomodulatory and anti-inflammatory effects through a combined “dispersion–astringency” mechanism. Meanwhile, the “middle energizer (zhongjiao) functional unit,” represented by ginger, jujube, and licorice, regulates gut microbiota composition through prebiotic-like effects. These two mechanisms act synergistically to achieve dual “immune–metabolic” regulation.

Furthermore, the intake of warm, diluted porridge rapidly provides fermentable substrates, promoting SCFA production and synergizing with pharmacological effects to improve metabolic homeostasis. This finding supports the classical TCM concept that “the spleen and stomach serve as the source of ying–wei transformation and generation,” as illustrated in Figure

5. 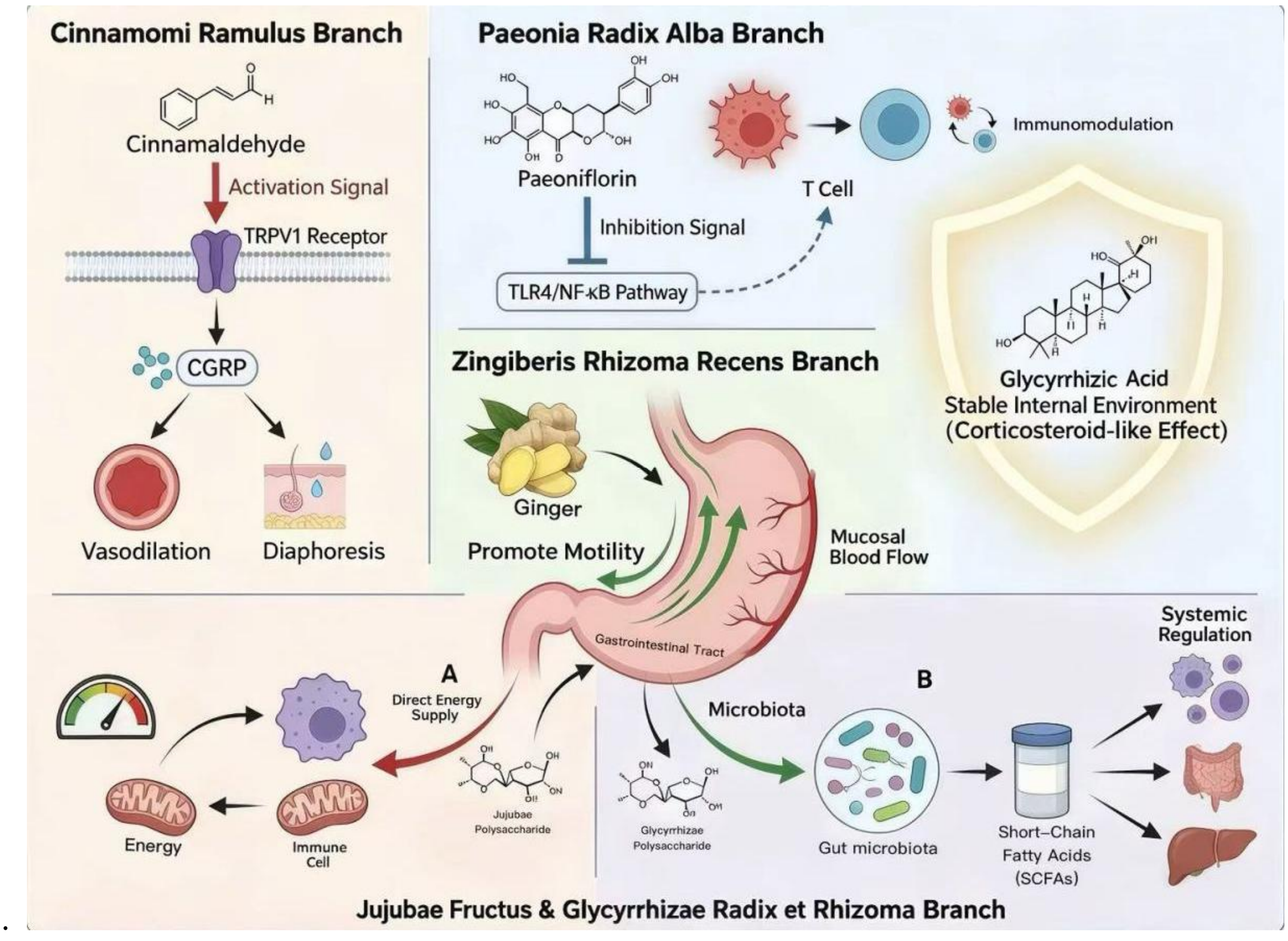

## Figure 5. The Compatibility–Metabolism–Microbiota Synergy Hypothesis

### Hypothesis 3: Dynamic Effector Substance Collection Hypothesis and Its Research Plan

The in vivo effect of Guizhi Decoction does not rely on a single component, but rather on a dynamic ensembnle composed of "prototype components (minor amount), metabolites (e.g., cinnamic acid and paeoniflorin derivatives), microbiota-transformed products (e.g., glycyrrhetinic acid)." This ensemble acts synergistically on core pathways such as cholinergic synapses, NF-κB, and the HPA axis through multi-target mechanisms, thereby exemplifying the characteristic of "Multi-Component-Multi-Target-Multi-Pathway" regulatory paradigm of TCM, as illustrated in Figure 6.

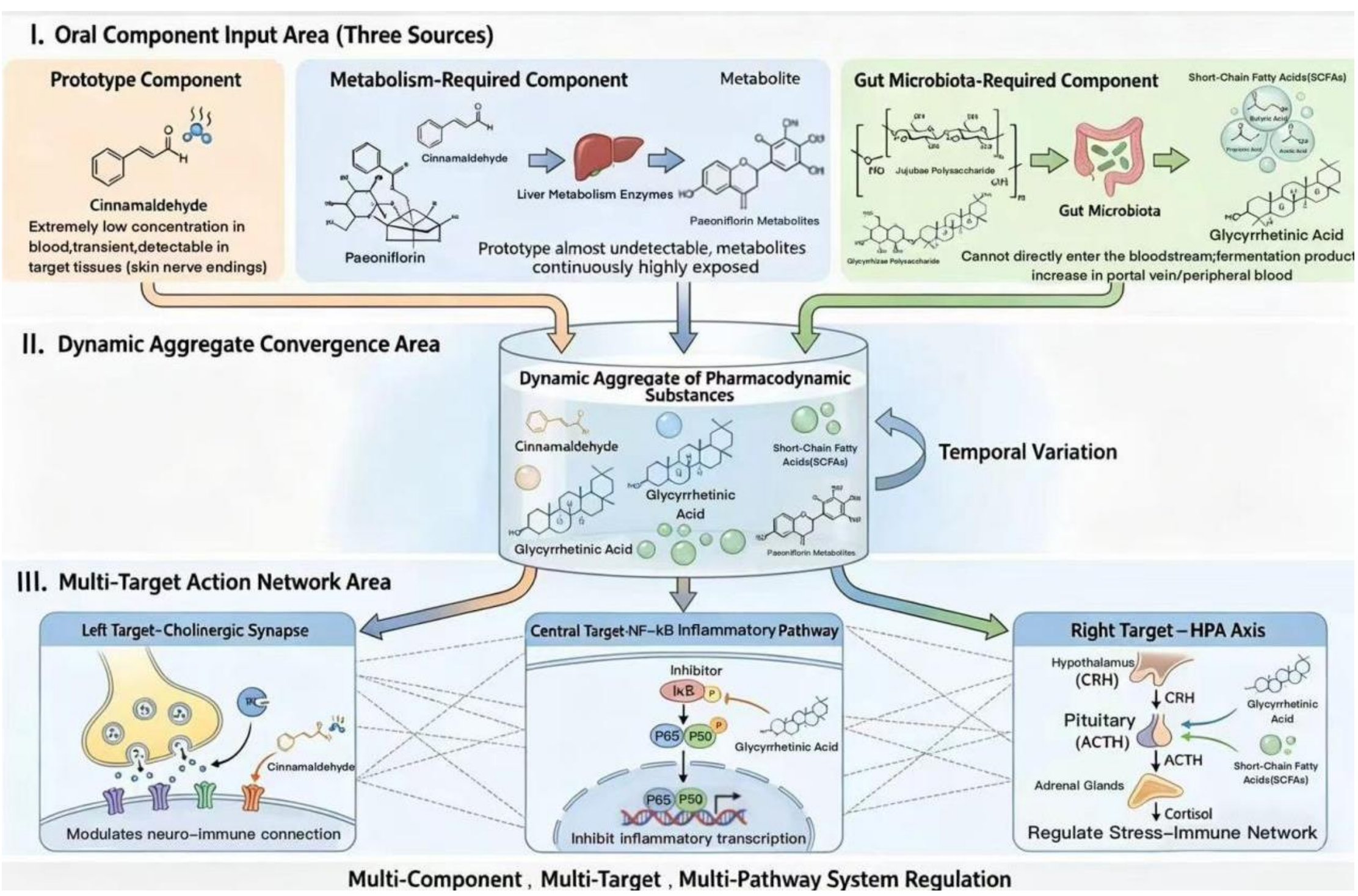

**Figure 6. The Dynamic Ensemble Hypothesis of Bioactive Substance Assemblages**

### 3.5 Multidimensional Comprehensive Performance Evaluation

#### 3.5.1 Double-Blind Scoring Experimental Design

The evaluation experiment employed a double-blind anonymous five-dimensional Likert 5-point scoring system to assess the quality of five AI-generated analytical reports on the modernization-oriented mechanistic interpretation of Guizhi Decoction. The evaluated reports included one multi-agent collaboratively generated report (the DeepTCM1.0 report) and four single large language model-generated reports (the DeepSeek V4 report, Doubao report, Kimi k2.6 report, and Qwen3.5 report).

To minimize subjective bias associated with evaluation by any single model, this study selected four representative large language models—DeepSeek, Doubao, Kimi, and Qwen—as independent evaluators. Each model independently scored every anonymous report according to a predefined standardized scoring rubric.

The scoring procedure strictly adhered to the double-blind principle. First, all five reports were anonymized by removing source-identifying information and assigning unified anonymous labels (R1–R5). The evaluation models assessed each report solely on the basis of textual content, without knowledge of its origin. After scoring was completed, the reports were unblinded for disclosure and comparative analysis. The entire evaluation process was fully traceable and verifiable.

### 3.5.2 Evaluation Dimension Design

Five core evaluation dimensions were defined in this study (Likert 5-point scale, 1 = very poor, 5 = excellent):

1. Completeness and Accuracy of Classical TCM Content: This dimension evaluates the completeness and correctness of classical TCM references, interpretations of pathogenesis, compatibility logic, and explanations of decoction and administration methods.

2. Comprehensiveness and Rigor of Modern Scientific Content: This dimension assesses the coverage and data reliability of modern scientific research elements, including molecular mechanisms, signaling pathways and targets, pharmacokinetics, and microbiota metabolism.

3. Interdisciplinary Integration Capability: This dimension evaluates the extent of cross-disciplinary linkage, synergistic reasoning, and depth of integration between TCM theoretical frameworks and modern scientific mechanisms.

4. Systematic Integrity and Logical Closure: This dimension assesses the structural framework of the report, logical coherence, completeness of argumentation, and absence of internal contradictions.

5. Standardization and Rigor of Academic Expression: This dimension evaluates the proper use of terminology, evidence annotation, hypothesis specification, disclosure of limitations, and the prudence and restraint of conclusions.

### 3.5.3 Evaluation and Comparative Framework

1. Evaluation Subjects

Multi-agent group: The DeepTCM1.0 multi-agent collaborative report.

Single large language model group: Reports independently generated by four single-model systems, DeepSeek V4, Doubao, Kimi k2.6, and Qwen 3.5.

2. Double-Blind Scoring Procedure

To minimize stochastic variability, positional bias, and output instability commonly observed in large language model evaluation, this study strictly followed a standardized double-blind evaluation workflow consisting of anonymization → repeated independent scoring → score aggregation → unblinded disclosure.

First, all five reports were anonymized by removing source-identifying information and assigning anonymous labels. Evaluators scored each report solely on the basis of textual content. Second, the four large language models served as independent evaluators, and each report was assessed five times under randomized presentation order, thereby eliminating order bias and reducing single-instance stochastic error. This yielded a total of 100 independent scoring assessments. Third, total scores were calculated through cumulative summation across the five evaluation dimensions. Finally, after scoring completion, report identities, report names, scores, and generation modes were disclosed, ensuring full traceability throughout the evaluation process.

To further ensure objectivity, the order of the five reports was randomly shuffled before each evaluation round to eliminate order bias; all evaluators operated under a unified standardized system prompt to constrain scoring behavior and enhance output consistency; each scoring session was conducted in a newly initialized dialogue context to eliminate historical context interference.

### 3.5.4 Validity Verification Metrics

To establish the reliability and validity of the repeated large-model-based evaluation framework, four complementary validation metrics were introduced.

1. Reliability Testing

(1) Inter-Evaluator Reliability: Krippendorff's α

Raw dimension-level scores (five dimensions) were used as input to calculate Krippendorff's alpha coefficient, evaluating scoring consistency among the four independent evaluators.

$$\alpha = 1 - \frac{D_o}{D_e}$$

where: $D_o$ is observed disagreement, and $D_e$ is expected disagreement by chance.

Interpretation criteria: α ≥ 0.7 means acceptable consistency. A ≥ 0.8 means excellent consistency

(2) Intra-Evaluator Reliability: Intraclass Correlation Coefficient (ICC)

The total scores obtained across five repeated scoring rounds for each evaluator were analyzed using the two-way random-effects absolute-agreement ICC(2,1) model.

$$ICC(2,1) = \frac{MS_{obj} - MS_{err}}{MS_{obj} + \frac{1}{k}(MS_{rater} - MS_{err})}$$

where: $MS_{obj}$ is mean square between reports; $MS_{err}$ is residual mean square; $MS_{rater}$ is mean square between evaluators; k is number of evaluators.

Interpretation criteria: ICC≥0.7 means acceptable reliability; ICC ⩾ 0.8 means good internal reliability.

2. Validity and Difference Testing

(1) Discriminative Validity: One-Way ANOVA and Effect Size η²

A one-way repeated-measures ANOVA was conducted using report identity (five reports) as the independent variable and all 100 scores as dependent observations.

$$F = \frac{MS_{between}}{MS_{within}}$$

where: $MS_{between}$ is between-group mean square; $MS_{within}$ is within-group mean square. Larger F values indicate stronger between-group discrimination.

Effect size was quantified using eta-squared:

$$\eta^2 = \frac{SS_{between}}{SS_{total}}$$

where: $SS_{between}$ is between-group sum of squares; $SS_{total}$ is total sum of squares.

Interpretation: a statistically significant ANOVA result (($P < 0.05$)) indicates effective discrimination between reports, while larger $\eta^2$ values indicate greater explanatory power.

(2) Intergroup Difference Testing: Mann–Whitney U Test

Given the relatively small sample size and non-normal distribution assumptions, Mann–Whitney U testing (Wilcoxon rank-sum test) was used to compare total scores between the multi-agent and single-model groups.

$$U_1 = n_1 n_2 + \frac{n_1(n_1+1)}{2} - T_1,\quad U_2 = n_1 n_2 + \frac{n_2(n_2+1)}{2} - T_2$$

where: $n_1$、$n_2$ are sample sizes; $T_1$、$T_2$ are rank sums. Smaller U values indicate stronger group differences.

The standardized Z statistic was computed as:

$$Z = \frac{U - \frac{n_1 n_2}{2}}{\sqrt{\frac{n_1 n_2 (n_1 + n_2 + 1)}{12}}}$$

(3) Effect Size Calculation

Effect size for rank-sum testing was calculated as:

$$r = \frac{\lfloor Z \rfloor}{\sqrt{N}}$$

where: N is total sample size.

Interpretation criteria: r⩾0.5 means large effect; 0.3⩽r<0.5 medium effect; r<0.3 small effect.

### 3.5.5 Visualization Methodology

Quantitative results were visualized using five-dimensional radial radar plots, generating five independent subfigures corresponding to: DeepTCM1.0, DeepSeek V4, Doubao, Kimi k2.6, Qwen3.5. All radar plots employed the same five evaluation axes: Completeness and accuracy of classical TCM content, Comprehensiveness and rigor of modern scientific analysis, Interdisciplinary integration capability, Logical systematicity and closed-loop coherence, Academic standardization and rigor. The scoring scale ranged from 1 to 5, strictly aligned with the Likert five-point standard.

Within each radar plot, dimension-wise mean scores across five repetitions were plotted for each evaluator. A combined mean contour representing the overall evaluator average was superimposed as a reference. Thus, each subplot contained five scoring curves.

The dispersion among evaluator contours visually reflected scoring consistency and stability, while radar coverage area, dimensional distribution, and deviation from the overall mean contour enabled intuitive cross-report comparison of multidimensional performance characteristics.

## 3.6 Results of Multi-Model Performance Evaluation

### 3.6.1 Reliability and Validity Analysis

Intra-evaluator reliability demonstrated strong internal consistency. ICC values for the four evaluators Doubao, DeepSeek, Kimi, Qwen are 0.867, 0.819, 0.789, 0.798.

All exceeded the acceptable threshold of ICC ≥ 0.7, indicating stable repeated scoring performance and controllable random error.

Inter-evaluator reliability was also excellent. Krippendorff's α coefficients across the five dimensions ranged from 0.842 to 0.867, with an overall total-score coefficient of 0.868. This exceeded the 0.8 threshold for excellent consistency, indicating strong evaluator agreement and minimal subjective bias.

One-way ANOVA revealed highly significant intergroup differences: F=66.969， P < 0.001, with effect size $\eta^2=0.738$. This indicates that 73.8% of total score variance could be explained by report quality differences, confirming excellent discriminative validity.

The Mann–Whitney U test further demonstrated highly significant differences between the multi-agent and single-model groups: U=191.00，Z=-5.403，P < 0.001, with rank-based effect size: r=0.5403. This reached the threshold for a large effect, confirming that the DeepTCM1.0 multi-agent framework significantly outperformed single general-purpose large language models, with highly robust statistical significance.

### 3.6.2 Radar Chart Visualization Results

Scores across the five evaluation dimensions for each model were normalized, and a composite radar chart comparing multi-model performance was generated (Figure 7). Subfigures a–e correspond to different models, respectively. The radial axis represents the score values (range: 1–5 points, Likert 5-point scale), while the dimensional axes sequentially represent completeness of classical TCM content, comprehensiveness of modern scientific content, interdisciplinary integration capability, systematic logical closure, and standardization of academic expression.

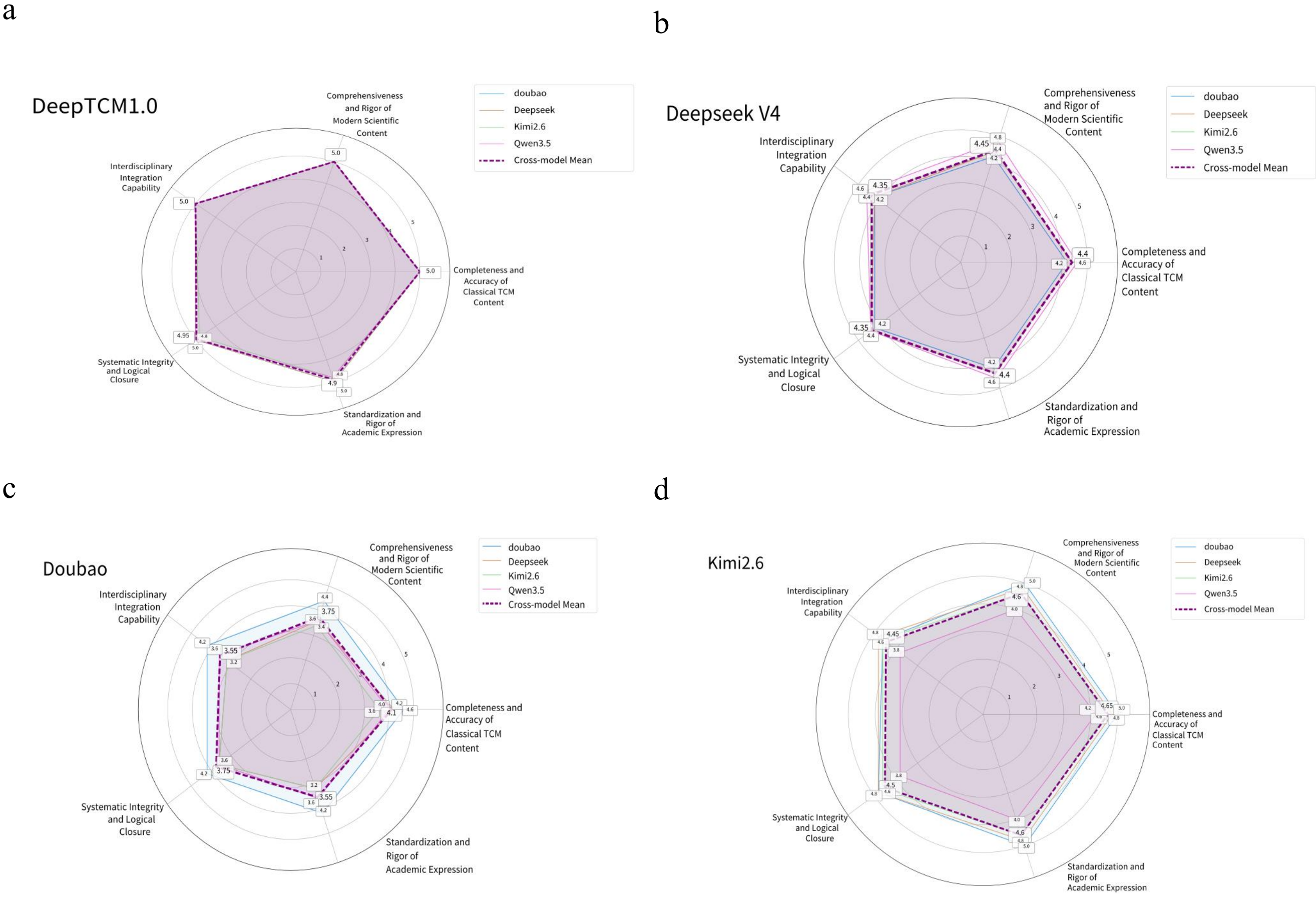

e

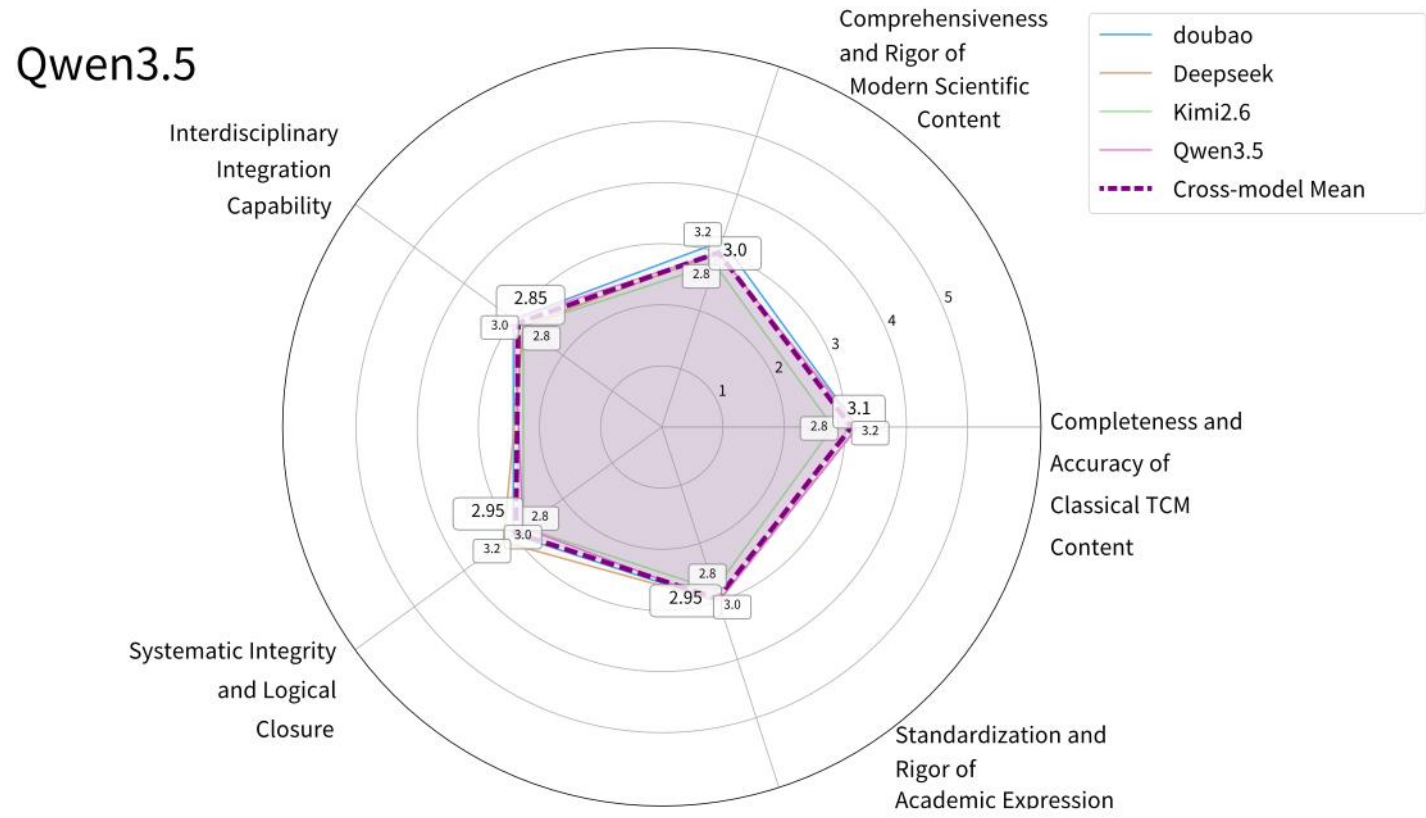

**Figure 7. Combined Radar Chart of Multi-Model Performance Comparison.**

**a: DeepTCM1.0 multi-agent model; b: DeepSeek V4 single model; c: Doubao single model; d: Kimi k2.6 single model; e: Qwen 3.5 single model.**

**Core findings:** As illustrated in the radar plots, DeepTCM1.0 achieved near-ceiling performance across all five evaluation dimensions, with scores ranging from 4.8 to 5.0, approaching the theoretical maximum. Notably, score contours from all evaluators were highly convergent, indicating exceptionally strong inter-evaluator agreement and highly stable performance across repeated assessments.

Among the single-model systems, Kimi k2.6 demonstrated relatively strong overall performance, maintaining high scores between 4.0 and 5.0, suggesting solid capability in both classical TCM interpretation and modern interdisciplinary reasoning. DeepSeek V4 exhibited stable upper-middle performance, with scores consistently distributed between 4.2 and 4.8, reflecting comparatively balanced but less comprehensive analytical depth.

In contrast, Doubao and Qwen3.5 showed substantially weaker performance, with most dimensional scores concentrated within the 2.8–4.2 range, generally falling below the overall mean reference contour. Their radar profiles also displayed greater dimensional fluctuation, indicating weaker consistency and less balanced cross-dimensional competence.

Collectively, the visualized results provide intuitive and compelling evidence that the DeepTCM1.0 multi-agent collaborative framework significantly outperforms single general-purpose large language models across all evaluation dimensions. Beyond achieving superior absolute scores, DeepTCM1.0 also demonstrates greater evaluator consistency, stronger dimensional balance, and enhanced robustness in interdisciplinary knowledge integration, further substantiating the effectiveness and methodological superiority of the proposed multi-agent architecture.

### 3.6.3 Summary of Scoring Statistics

Across the five reports under evaluation, four independent evaluator models conducted five repeated scoring rounds, and the aggregated mean scores for each dimension as well as total score statistics are presented in Table 2 (see Supplementary Table 1 for detailed scoring records). Quantitative results showed that the DeepTCM1.0 multi-agent framework achieved the highest scores across all five evaluation dimensions, with its mean total score significantly exceeding those of all single-model systems.

Clear hierarchical differences were observed among the four general-purpose large language models. Kimi k2.6 achieved the highest overall mean score among the single-model baselines, followed by DeepSeek V4, whereas Doubao and Qwen3.5 obtained comparatively lower overall scores. Across dimensions including classical TCM interpretation, modern scientific reasoning, and interdisciplinary integration, varying degrees of performance divergence were evident among the models.

These quantitative results objectively reflect the capability differences of different models in the task of mechanistic interpretation of traditional Chinese medicine formulae. Together with the radar visualization results, they further demonstrate that the DeepTCM1.0 system—characterized by multi-expert collaboration, three-round iterative reasoning, and critic-based quality control—effectively compensates for the limitations of single general-purpose large language models in the TCM domain.

Accordingly, this framework is better suited to the inherent complexity of TCM formula research, which involves multi-component interactions, multi-target regulatory mechanisms, interdisciplinary knowledge integration, and strong theoretical dependence on classical TCM foundations.

**Table 2. Mean Scores Across Five Evaluation Dimensions and Total Scores for Each Model**

| **Report Name** | **Completeness of classical TCM content** | **Comprehensiveness of modern scientific content** | **Interdisciplinary integration capability** | **Systematic logical closure** | **Standardization of academic expression** | **Mean ± SD** |
|---|---|---|---|---|---|---|
| DeepTCM1.0 | 5.00 | 5.00 | 5.00 | 4.95 | 4.90 | 24.85±0.19 |
| DeepSeek V4 | 4.40 | 4.45 | 4.35 | 4.35 | 4.40 | 21.95±0.82 |

| Report Name | Completeness of classical TCM content | Comprehensiveness of modern scientific content | Interdisciplinary integration capability | Systematic logical closure | Standardization of academic expression | Mean ± SD |
| --- | --- | --- | --- | --- | --- | --- |
| Doubao | 4.10 | 3.75 | 3.55 | 3.75 | 3.55 | 18.70±2.02 |
| Kimi k2.6 | 4.65 | 4.6 | 4.45 | 4.50 | 4.60 | 22.80±2.08 |
| Qwen3.5 | 3.10 | 3.00 | 2.85 | 2..95 | 2..95 | 14.85±0.62 |

## 4 Discussion

The multi-expert agent collaboration framework, DeepTCM 1.0, based on a general large model proposed in this study successfully achieved a multi-dimensional analysis of the mechanism underlying Chinese medicine formulas, using Guizhi Decoction as a representative case. The results show that this framework can effectively integrate cross-disciplinary knowledge, construct logically coherent mechanistic hypotheses, and demonstrate distinct advantages in aligning with TCM theoretical principles.

### 4.1 Comparison with Existing Technical Paths

Compared with existing technologies such as data mining, network pharmacology, and deep learning models, the core value of DeepTCM 1.0 can be summarized in three aspects.

#### 4.1.1 Completeness of Cross-Disciplinary Knowledge Integration

Conventional network pharmacology research is often constrained by separate databases and static network analysis, making it difficult to achieve deep fusion of multi-disciplinary knowledge. Additionally, these approaches are associated with time-consuming data organization and a disconnec between analytical outputs and theoretical interpretation [26, 34, 35, 41]. In contrast, the present framework employed 11 expert agents with specialized roles and dynamic collaboration, integrating knowledge from classical TCM theory, clinical practice, modern pharmacology, and systems biology. This resulted in the construction of a comprehensive knowledge chain spanning “classical theory–microscopic mechanisms–clinical application,” thereby effectively addressing the issue of inefficient cross-domain knowledge integration identified in the Introduction.

#### 4.1.2 Logical Closed-Loop Nature of Mechanism Analysis

Existing deep learning models in TCM research universally suffer from the limitations of "black box prediction"—namely, opaque reasoning processes, lack of traceability in conclusions, disconnection from core TCM theories, and superficial integration of cross-disciplinary knowledge [36, 37, 38, 42]. All hypotheses in this framework follow the closed-loop logic of "Classic Traceability - Evidence Support - Logical Deduction." The "Three-Layer Network Harmonization Hypothesis" not only originates from the Ying-Wei theory derived from Shanghan Lun, but also receives analytical support from multiple disciplines such as network pharmacology target prediction and metabolomics microbiota analysis, ensuring that every conclusion is traceable to its source. Each conclusion was thus rendered traceable, and could be further validated through pharmacokinetic–pharmacodynamic (PK–PD) modeling to simulate the in vivo dynamics of bioactive substances. The full-process quality control conducted by the Critic further ensures the integrity of the argumentation logic and the reliability of the evidence, effectively addressing the issue of discontinuity in knowledge integration.

#### 4.1.3 Precision of TCM Theory Adaptability

Current research predominantly relies on pharmacological databases, with analytical outputs expressed in the language of modern molecular biology. This often leads to a lack of theoretical grounding in TCM, resulting in mechanistic interpretations that deviate from TCM paradigms and neglect its holistic perspective [33, 39, 40, 43]. In contrast, the proposed framework consistently prioritized classical TCM theory. It emphasized key concepts such as “taking warm porridge and maintaining warmth after decoction administration” (cuozhou wenfu) and the functional role of the “middle energizer (zhongjiao),” while strictly adhering to the classical 1:1 ratio of Guizhi (Cinnamomi Ramulus) to Shaoyao (Paeoniae Radix). This approach prevented the partial misinterpretation of TCM theory by modern analytical techniques. Furthermore, the design of quantitative indicators, such as the “ying–wei harmony index” and the dynamic criterion of “self-regulation of yin–yang balance,” enabled the simulation and quantification of classical theoretical constructs, highlighting the unique advantage of the multi-agent framework in achieving precise technical adaptation to TCM principles.

### 4.2 Implications for TCM Methodology

The primary innovation of this study lies in exploring a new path for TCM mechanism analysis that is lightweight, low-cost, and reproducible. Without relying on additional database construction or model fine-tuning, it significantly lowers the application threshold of AI technology in TCM research. Through refined role prompt engineering, it achieves the "role transformation" of general large models into TCM domain experts. Moreover, all prompts and process designs are shareable, facilitating other researchers to migrate them to studies on different classic formulas. This technology is expected to promote the transformation of TCM research from experience-based knowledge inheritance to intelligence-driven scientific innovation.

## 4.3 Limitations and Future Directions

This study still has certain limitations. First, the framewrok relies entirely on the pre-trained knowledge of general large models, which may result in incomplete coverage of domain specific TCM knowledge, particularly wotj respect to rare classic texts and region-specific medical traditions. Second, the analysis depth of the agents is affected by the quality of prompt design, and the optimization of prompt engineering requires iterative exploration. Third, the proposed mechanism hypotheses still require experimental verification, as the current findings are primarily derived from knowledge-based inference.

Future research can be advanced in the following directions. First, Retrieval-Augmented Generation (RAG) technology could be incorporated to dynamically expand domain knowledge coverage without model fine-tuning, for example by integrating real-time updated databases like TCMSP [44] and SymMap [45]. Second, automatic prompt optimization algorithms could be developed to iteratively optimize role prompts based on historical discussion outcomes. Third, collaborate with experimental research teams should be pursued to validate key pathways (e.g., NF-κB, TRPV1) and critical biomarkers (e.g., SCFAs, TNF-α/IL-10 ratio). Fourth, the framework could be extend to other classic formulas, such as Mahuang Decoction and Xiao Chaihu Decoction to test its generalizability and transferability.

## 5 Conclusion

In this study, a multi-expert agent collaboration framework, DeepTCM 1.0, based on general large language model is developed. Using Guizhi Decoction as a representative case, a multi-dimensional mechanism analysis of TCM formulae was successfully achieved. Through a three-layer collaboration architecture and a three-round iterative discussion mechanism, this framework effectively integrated multi-disciplinary knowledge such as TCM classics, formula compatibility, and modern pharmacology, thereby generating innovative mechanism hypotheses with a logical closed loop. Compared with existing technologies, this framework demonstrates significant advantages in cross-disciplinary knowledge integration, TCM theory adaptability and interpretability,. It provides a lightweight and reproducible paradigm for the analysis of complex TCM systems. Future work will focus on experimental validation and framework extension to further promote its application in the modernization of TCM research.

## Availability of data and materials

Code and data during the current study is available from the corresponding author upon reasonable request.

## Funding Declaration

The authors received no specific funding for this work.

**Appendix**

**Table A1. Detailed Scores of Each Dimension and Total Score from 5 Rounds of Repeated Scoring**

| Report Name | Evaluation Model | Completeness of Classical TCM Content | Comprehensiveness of Modern Scientific Content | Interdisciplinary Integration Capability | Systematic Logical Closure | Standardization of Academic Expression | Total Score |
|---|---|---|---|---|---|---|---|
| Deep TCM 1.0 | Doubao | 5 | 5 | 5 | 5 | 5 | 25 |
| Deep TCM 1.0 | DeepSeek | 5 | 5 | 5 | 5 | 5 | 25 |
| Deep TCM 1.0 | Kimi2.6 | 5 | 5 | 4.8 | 4.8 | 4.8 | 24.6 |
| Deep TCM | Qwen3.5 | 5 | 5 | 5 | 5 | 4.8 | 24.8 |

| Report Name | Evaluation Model | Completeness of Classical TCM Content | Comprehensiveness of Modern Scientific Content | Interdisciplinary Integration Capability | Systematic Logical Closure | Standardization of Academic Expression | Total Score |
|---|---|---|---|---|---|---|---|
| 1.0 | | | | | | | |
| DeepSeek V4 | Doubao | 4.2 | 4.2 | 4.2 | 4.2 | 4.2 | 21 |
| DeepSeek V4 | DeepSeek | 4.4 | 4.4 | 4.2 | 4.4 | 4.4 | 21.8 |
| DeepSeek V4 | Kimi2.6 | 4.4 | 4.4 | 4.4 | 4.4 | 4.4 | 22.0 |
| DeepSeek V4 | Qwen3.5 | 4.6 | 4.8 | 4.6 | 4.4 | 4.6 | 23.0 |
| Doubao | Doubao | 4.6 | 4.4 | 4.2 | 4.2 | 4.2 | 21.6 |
| Doubao | DeepSeek | 4.2 | 3.6 | 3.2 | 3.6 | 3.2 | 17.8 |
| Doubao | Kimi2.6 | 3.6 | 3.4 | 3.2 | 3.6 | 3.2 | 17 |
| Doub | Qwen3. | 4.0 | 3.6 | 3.6 | 3.6 | 3.6 | 18.4 |

| Report Name | Evaluation Model | Completeness of Classical TCM Content | Comprehensiveness of Modern Scientific Content | Interdisciplinary Integration Capability | Systematic Logical Closure | Standardization of Academic Expression | Total Score |
|---|---|---|---|---|---|---|---|
| ao | 5 | | | | | | |
| Kimi k2.6 | Doubao | 5 | 5 | 4.6 | 4.8 | 5 | 24.4 |
| Kimi k2.6 | DeepSeek | 4.8 | 4.8 | 4.8 | 4.8 | 4.8 | 24 |
| Kimi k2.6 | Kimi2.6 | 4.6 | 4.6 | 4.6 | 4.6 | 4.6 | 23 |
| Kimi k2.6 | Qwen3.5 | 4.2 | 4. | 3.8 | 3.8 | 4.0 | 19.8 |
| Qwen 3.5 | Doubao | 3.2 | 3.2 | 3 | 3 | 3 | 15.4 |
| Qwen 3.5 | DeepSeek | 3.2 | 3 | 2.8 | 3.2 | 3 | 15.2 |
| Qwen 3.5 | Kimi2.6 | 2.8 | 2.8 | 2.8 | 2.8 | 2.8 | 14 |
| Qwen 3.5 | Qwen3.5 | 3.2 | 3 | 2.8 | 2.8 | 3 | 14.8 |